\documentclass[10pt,journal,compsoc]{IEEEtran}
\usepackage{amsmath,amsfonts}
\usepackage{algorithmic}
\usepackage{array}
\usepackage[caption=false,font=scriptsize,labelfont=sf,textfont=sf]{subfig}
\ifCLASSOPTIONcompsoc
\usepackage[nocompress]{cite}
\else
\usepackage{cite}
\fi
 
\usepackage{textcomp}
\usepackage{stfloats}
\usepackage{url}
\usepackage{verbatim}
\usepackage{graphicx}
\usepackage{balance}
\usepackage{stfloats}
\usepackage{dsfont} 

\usepackage{multirow}
\usepackage{booktabs}
\usepackage{svg} 
\usepackage{makecell}
\usepackage{graphicx}
\usepackage{epsfig}
\usepackage{amsmath}
\usepackage{amssymb}
\usepackage{algorithm}
\usepackage{algorithmic}

\begin{document}

\title{Do Fairness Interventions Come at the Cost of Privacy: Evaluations for Binary Classifiers}
\author{~Huan~Tian,
        ~Guangsheng~Zhang,
        ~Bo~Liu,
        ~Tianqing~Zhu*,
        ~Ming~Ding,
        ~Wanlei~Zhou
\IEEEcompsocitemizethanks{\IEEEcompsocthanksitem Tianqing Zhu is the corresponding author. H. Tian, G. Zhang, B. Liu are with Australian Artificial Intelligence Institute and the School of Computer Science, University of Technology Sydney, Australia. Email: \{Huan.Tian, Guangsheng.Zhang\}@student.uts.edu.au, Bo.Liu@uts.edu.au. Ming Ding is with Data61, CSIRO, Australia. Email: Ming.ding@data61.csiro.au. T. Zhu and W. Zhou are with City University of Macau, Macao. Email: \{tqzhu, wlzhou\}@cityu.edu.mo.\protect\\
}
}

\IEEEtitleabstractindextext{%
\begin{abstract}
While in-processing fairness approaches show promise in mitigating biased predictions, 
their potential impact on privacy leakage remains under-explored.
We aim to address this gap by assessing the privacy risks of fairness-enhanced binary classifiers via membership inference attacks (MIAs) and attribute inference attacks (AIAs).
Surprisingly, 
our results reveal that enhancing fairness does not necessarily lead to privacy compromises.
For example, these fairness interventions exhibit increased resilience against MIAs and AIAs.
This is because fairness interventions tend to remove sensitive information among extracted features and reduce confidence scores for the majority of training data for fairer predictions.
However, during the evaluations, 
we uncover a potential threat mechanism that exploits prediction discrepancies between fair and biased models, leading to advanced attack results for both MIAs and AIAs.
This mechanism reveals potent vulnerabilities of fair models and poses significant privacy risks of current fairness methods. 
Extensive experiments across multiple datasets, attack methods, and representative fairness approaches confirm our findings and demonstrate the efficacy of the uncovered mechanism.
Our study exposes the under-explored privacy threats in fairness studies, 
advocating for thorough evaluations of potential security vulnerabilities before model deployments.
\end{abstract}

}
\maketitle

\section{Introduction}
In recent years, 
there have been remarkable advancements in various fields thanks to 
large models such as the GPT models~\cite{brown2020language} and the Segment Anything Model~\cite{kirillov2023segment}.
These models have proven to be highly effective,
but their success heavily relies on extensive training data, 
which often contains biased data distributions. 
This raises concerns about algorithmic fairness,
where the resulting trained models (\textit{biased models}) may exhibit discriminative performances for certain demographic subgroups~\cite{mehrabi2021survey}.
To address the issue, 
previous studies have proposed in-processing methods that modify the learning algorithm to remove bias during model training.
After applying these fairness interventions, 
the obtained fairness-enhanced models (\textit{fair models}) can provide more equitable performance across subgroups, 
thus mitigating unfairness predictions. 
However, 
despite promising to enhance fairness, 
their potential impact on privacy leakage remains under-explored.

One pioneering work by Chang and Shokri~\cite{chang2021privacy} has explored the privacy implications of fairness interventions through the lens of membership inference attacks (MIAs). With the model's predictions, MIAs aim to infer whether a given sample was part of the training data (sample membership). These attacks are widely used to assess privacy risks in models deployed via Machine Learning as a Service (MLaaS) ~\cite{shokri2017membership}. Building on this, the authors first train biased models and then apply fairness interventions to obtain fair models. 
They perform MIAs on both the biased and fair models, comparing the attack results before and after the interventions. 
Their findings reveal that fairness interventions improve the effectiveness of MIAs, suggesting a potential trade-off between achieving model fairness and model privacy.

Although the study by Chang and Shokri~\cite{chang2021privacy} offers valuable insights, it has limitations. 
Firstly, the evaluations primarily focused on decision tree models. Although they briefly explored simple convolutional neural networks (CNNs), their evaluation was limited to a single synthetic dataset. This leaves open questions regarding the privacy risks of fairness interventions for neural networks in real-world datasets, such as binary classifiers, which are prevalent in fairness studies.
Secondly, the study adopted only one type of attack--score-based membership inference attacks (MIAs)--for evaluations. While this approach provides insights, it might not fully characterize the privacy impact of fairness interventions.
Given that both fairness and privacy are crucial aspects of model trustworthiness, conducting thorough evaluations of these interventions is crucial.
To address these gaps, our work considers multiple attacks, including both membership inference attacks (MIAs) and attribute inference attacks (AIAs), to evaluate fair binary classifiers, thereby comprehensively assessing the associated privacy risks.

\begin{figure*}[t]
    \centering
    \subfloat[MIAs \label{fig:intro_mia}]{
        \includegraphics[width=0.23\textwidth]{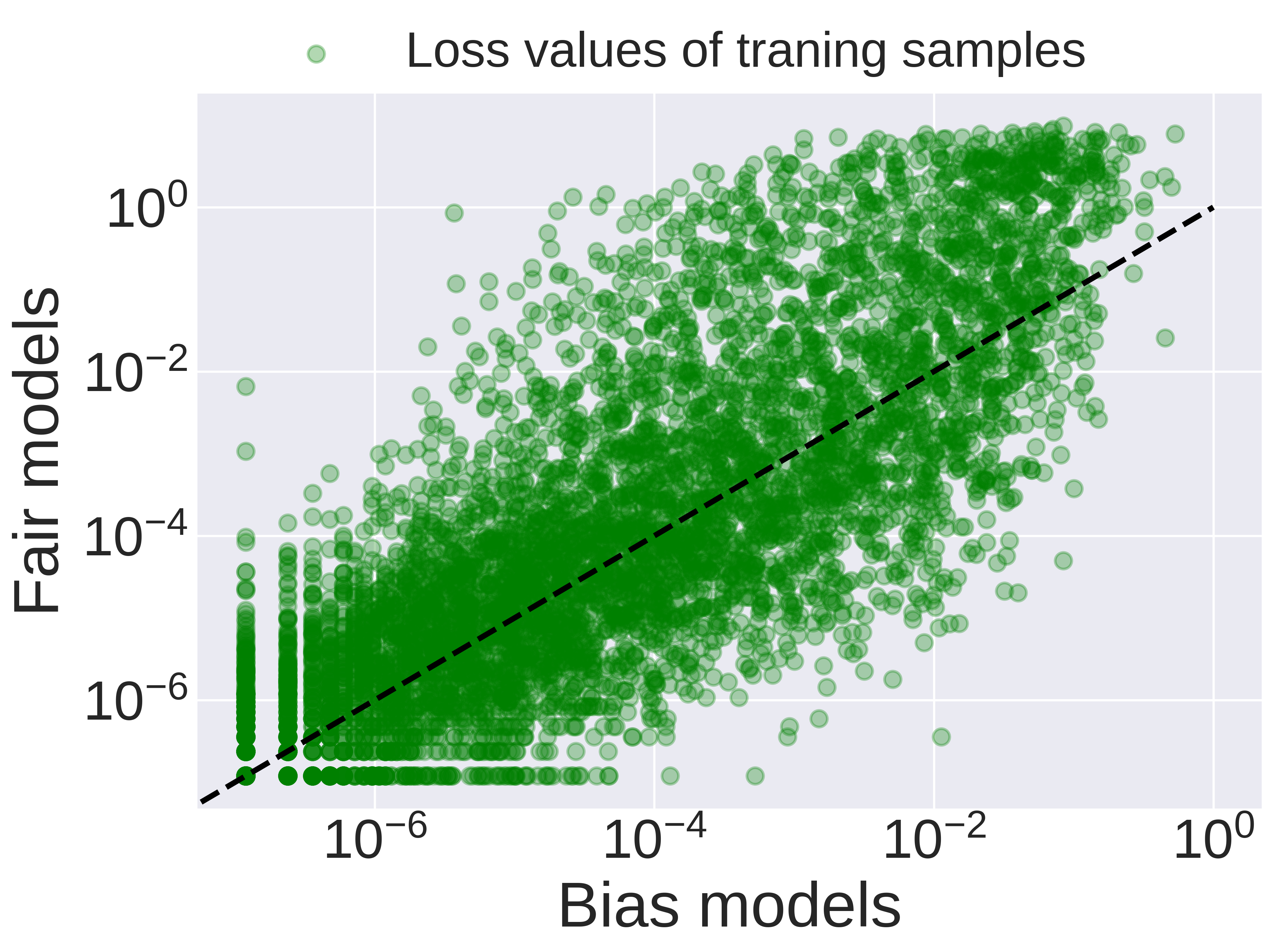}
        \includegraphics[width=0.23\textwidth]{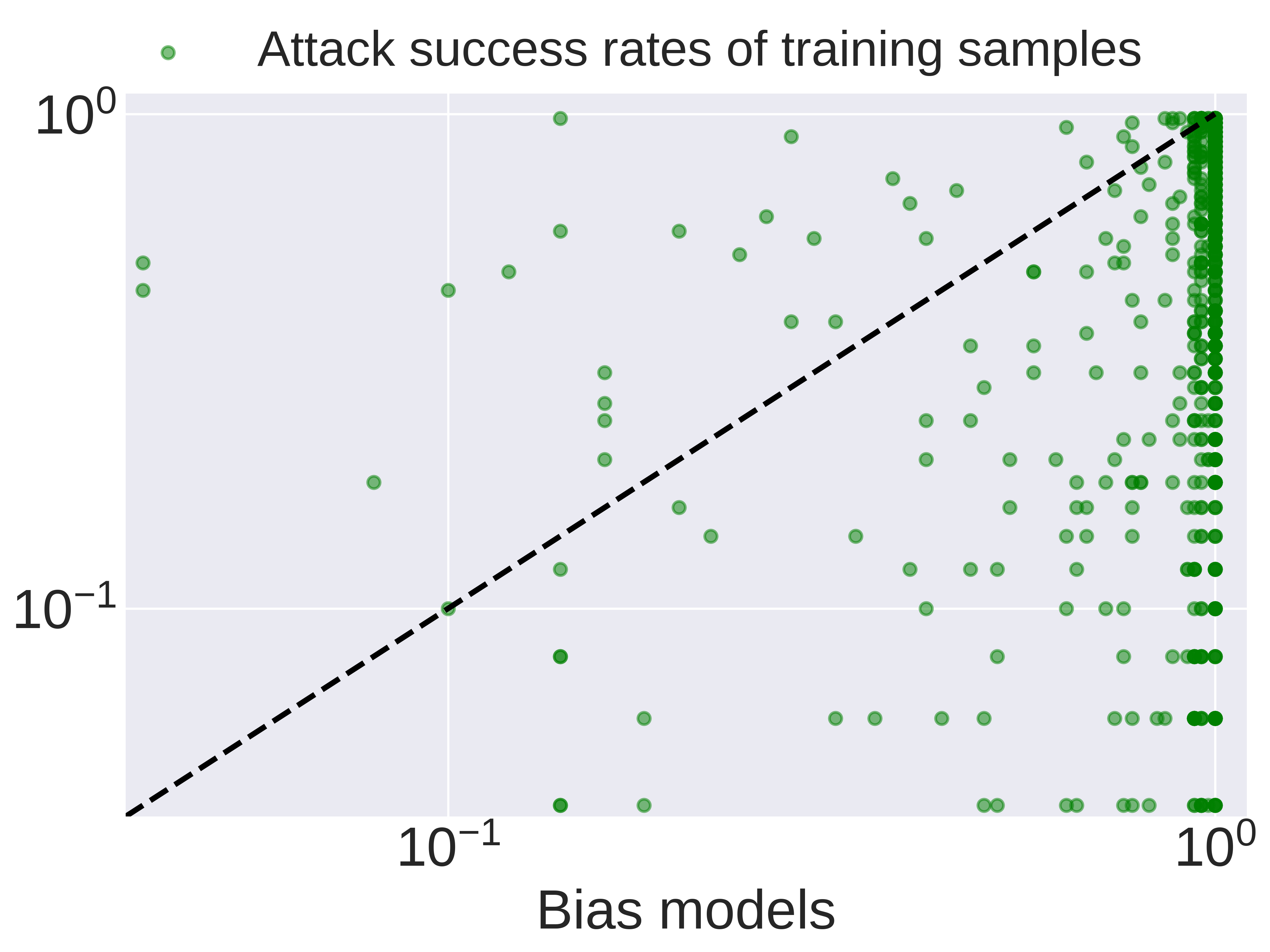}
    }
    \subfloat[AIAs \label{fig:intro_aia}]{
        \includegraphics[width=0.23\textwidth]{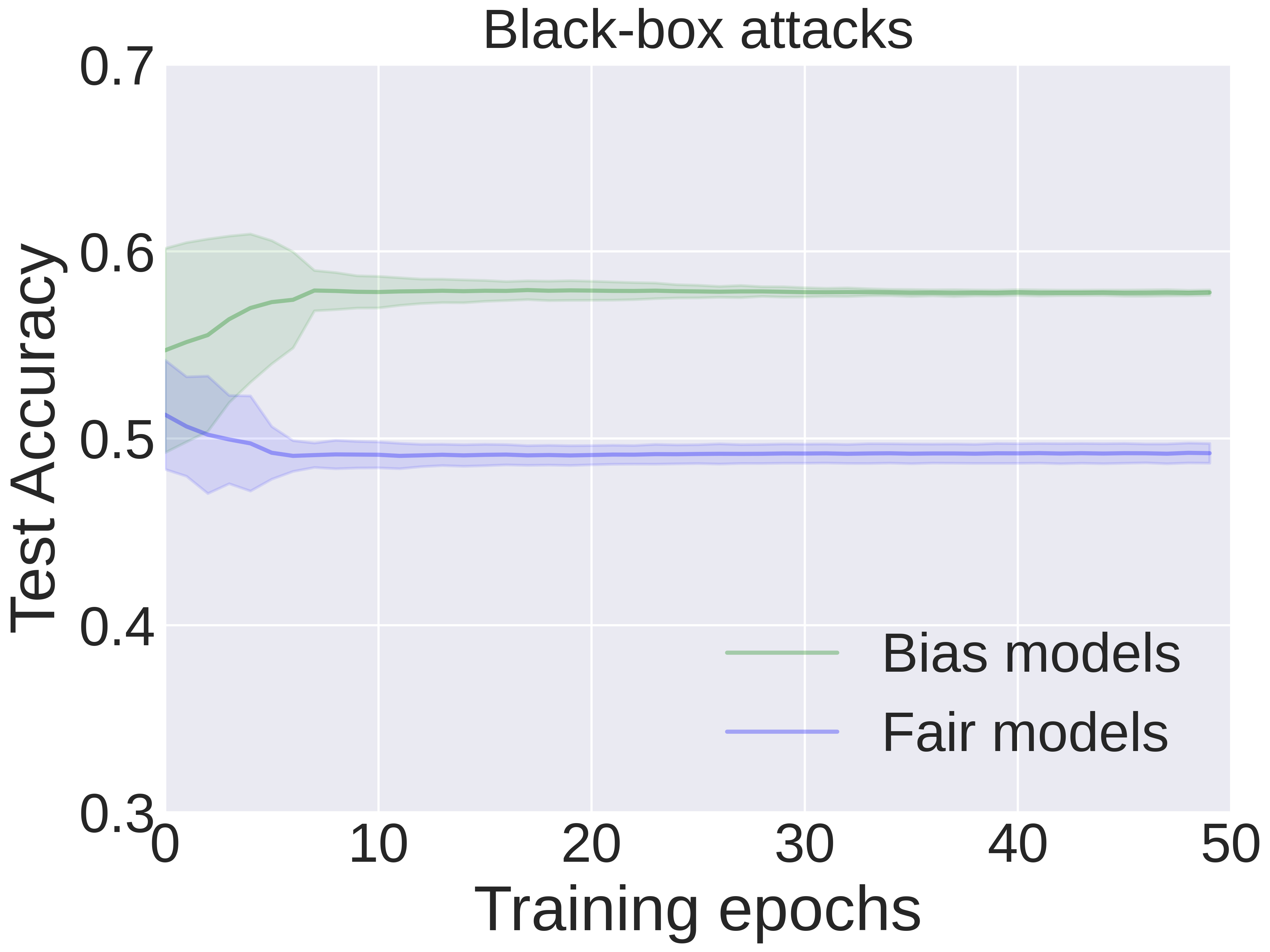}
        \includegraphics[width=0.23\textwidth]{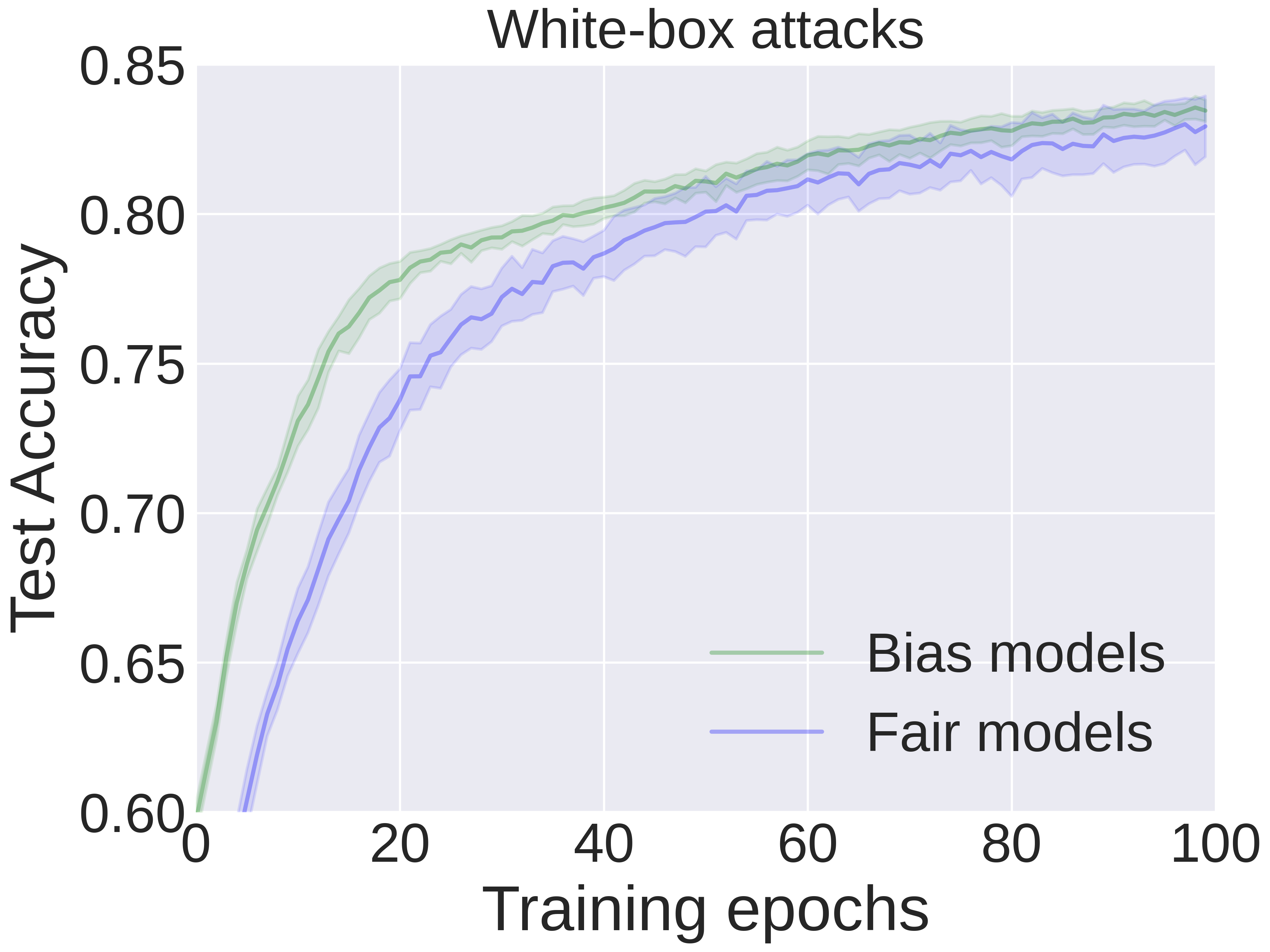}
    }
    \caption{Fairness interventions increase loss values for the majority training data, leading to diminished attack successful rates for MIAs (Figure~\ref{fig:intro_mia}). Meanwhile, compared to biased models, fair models show more resilience to AIAs under both Black and White-box settings (Figure~\ref{fig:intro_aia}).}
    \label{fig:intro}
\end{figure*}

Specifically, we evaluate the privacy of fair binary classifiers via membership inference attacks (MIAs) and attribute inference attacks (AIAs). While MIAs aim to infer given sample membership information, AIAs attempt to infer sensitive information about the sample, \textit{e.g.}, male or female for the gender attribute. To conduct thorough evaluations, we consider different attack methods for both MIAs and AIAs.
Surprisingly, our evaluation results show that \textbf{fair models show more resilience to current attacks than their biased counterparts.}
Figure~\ref{fig:intro} presents the MIAs and AIAs results on fair and biased models over $100$ runs. 
For MIAs, we report the per-sample loss values and the attack success rate (Figure~\ref{fig:intro_mia}) before and after applying fairness interventions--biased vs fair models. 
The plots show increased loss values yet decreased attack success rates for most data points with fair models. 
This indicates that these interventions can lead to less successful attacks with existing attack approaches.
For AIAs, we illustrate the test accuracy for both white and black-box AIAs (Figure~\ref{fig:intro_aia}). The plots show decreased attack results after fairness interventions, indicating inferior attack performance.

Through further analyses, we find that fairness interventions reduce sensitive information among the extracted features and confidence scores for the majority of training data, leading to fairer predictions. These reductions lead to more challenging attacks for both AIAs and MIAs, as the attacks have less exploitable information to leverage.
Meanwhile, our experiments reveal that the existing attack methods, which are primarily designed for multi-class scenarios, become less effective when applied to binary classification tasks. This inefficacy stems from the binary outputs, which cause the attack models to degrade into simple threshold-based decisions.
The degradation incurs substantial performance trade-offs. For example, in the case of MIAs, while effective at recognizing member data, the attack models struggle with non-member data. The phenomenon is particularly pronounced for ``hard examples''---samples where the predictions are similar across groups.

Before concluding that fairness interventions are privacy-friendly to binary classifiers,  
we further identify a potential threat that could enable more effective attacks. During the evaluation, we observe divergent prediction behaviors for different data groups after applying fairness interventions.
Specifically, the prediction scores typically increase for the majority of the training data and decrease for the minority training data. In contrast, the scores for non-member data conform to a normal distribution. This disparity creates pronounced prediction gaps between the data groups.
However, if adversaries exploit these widened gaps between groups, it might enable more successful attacks, thereby posing substantial privacy threats to fair binary classifiers.

Inspired by these observations, we introduce two new attack methods: Fairness Discrepancy-based Membership Inference Attacks (FD-MIAs) and Fairness Discrepancy-based Attribute Inference Attacks (FD-AIAs). These methods exploit the prediction gaps between the original biased models and the fair models more effectively.
The key is to leverage model fairness disparities.
Moreover, we demonstrate that FD-MIAs and FD-AIAs can be integrated with existing attack methods, such as score-based~\cite{LWHSZBCFZ22} and reference-based attacks~\cite{carlini2022membership}. This integration delivers advanced attack performance and poses real privacy threats to fair binary classifiers.

In the experiments, we conduct comprehensive evaluations across three datasets, 
with up to six attack methods and five in-processing fairness approaches. 
This amounts to $128$ different settings and more than $400$ distinct models.
The results consistently validate our findings and the identified threat.
Our study reveals that \textbf{fairness interventions can introduce new threats to model privacy}, 
advocating a more comprehensive examination of their potential security defects before deployment. 
Our main contributions are as follows:
\begin{itemize}
    \item To the best of our knowledge, this is the first work to comprehensively study the impact of fairness interventions on privacy through the lens of MIAs and AIAs, targeting deep classifiers with real-world datasets. 
    \item  We reveal that fairness interventions do not compromise model privacy with \textit{existing} attack methods, primarily due to their limited efficacy in attacking binary classifiers.
    \item  We identify a previously unexamined vulnerability and propose two novel attack methods, FD-MIA and FD-AIA, which pose real threats to model privacy by exploiting prediction gaps between biased and fair models. These methods can be integrated into existing attack frameworks.
    \item Extensive experiments on three datasets have confirmed our observations and demonstrated the efficacy of the proposed methods.
\end{itemize}

A preliminary version of this research was presented at IJCAI 2024 in~\cite{Huantian2024}. While the conference paper introduced the concept of privacy evaluations for fair models using MIAs, this manuscript significantly expands the scope and depth of our investigation. We broaden the privacy risk assessment by incorporating both MIAs and AIAs, providing a more comprehensive evaluation of fair binary classifiers. Furthermore, we extend the attack mechanism originally developed for MIAs to the domain of AIAs, resulting in novel and more potent attack strategies.

The organization of the paper is as follows: Section 2 reviews related work on fairness and privacy attacks. Section 3 outlines the preliminaries and evaluation metrics. In Section 4, we assess model privacy using membership inference attacks, while Section 5 extends this evaluation to attribute inference attacks. Building on these findings, Section 6 introduces our enhanced attack mechanism (FD-MIA and FD-AIA), which exploits prediction discrepancies between biased and fair models. Section 7 presents experimental evaluations, and Section 8 discusses potential mitigations and future research directions. Section 9 provides a broader discussion of our proposed methods. Sections 10 and 11 discuss the limitations and provide conclusions, respectively.

\section{Related work}
\subsection{Algorithmic fairness}
Given biased models, fairness methods aim to ensure consistent prediction performance across subgroups. According to the method modification phases, fairness studies generally fall into three categories: pre-processing, in-processing, and post-processing approaches. For deep classification models, studies usually adopt in-processing methods as they deliver fair results efficiently. Widely adopted methods involve the introduction of fair constraints, adversarial training, or mixup augmentation operations.

Fair constraint methods \cite{Zemel:2013wz,Manisha2020FNNCAF,Xu_2021_CVPR,NEURIPS2021_fc2e6a44,Tang_2023_CVPR,truong2023fredom,cruz2023fairgbm,jung2023reweighting} introduce additional constraints based on the fairness metrics. They formulate the problem as optimization issues. Initially proposed in \cite{Zemel:2013wz}, subsequent studies have developed the method with diverse settings such as proposing different constraints \cite{xu2020BeRobustBe,Xu_2021_CVPR,guo2023TNNLS} or training schemes \cite{Manisha2020FNNCAF}. Later, adversarial training methods have been proposed~\cite{Kim_2019_CVPR,madras2018learning,Zhu_2021_ICCV,zafar2019fairness,Park2021LearningDR}. These methods require additional predictions for sensitive attributes and update gradients reversely to remove sensitive information from extracted features. 
The operation leads to more similar representations across subgroups, 
contributing to fairer predictions. More recently, studies aim to learn ``neutral'' representations using mixup augmentation operations \cite{mroueh2021fair,du2021fairness} or contrastive learning~\cite{Park_2022_CVPR,Wang_2022_CVPR,zhang2023fairnessaware,qi2022fairvfl}. These methods either interpolate inputs or modify features to pursue fair representations. 
Other fairness methods include data operations such as balancing the data with synthetic data generation~\cite{hwang2020UnsupervisedImagetoImageTranslation,joo2020GenderSlopesCounterfactual,ramaswamy2021fair}, data sampling strategies \cite{roh2021fairbatch} ~\cite{Khalili2021FairSS} or data re-weighting strategies~\cite{zhao2020maintaining,gong2021mitigating} to enforce fairness. Others concentrate on different settings, such as semi-supervised learning~\cite{Zhang_2020,zhang2023fairnessaware,wei2023balanced}, multi-attribute protections~\cite{zafar2019fairness,huanmulti24,Huan2024multi}, or enforcing fairness without demographics~\cite{chai2022self}. 
In the experiments, we evaluate model privacy considering multiple fairness methods, delivering comprehensive evaluations.

\subsection{Membership inference attacks}
\textbf{Membership inference attacks} aim to determine whether a given data sample was in the target model's training dataset or not~\cite{shokri2017membership}.
A number of attacks leverage the target model's direct output as inputs to train the attack models and infer the membership of queried samples. For example, various studies~\cite{shokri2017membership,salemMLLeaksModelData2019,LWHSZBCFZ22} utilize the confidence scores as input, while others~\cite{yeomPrivacyRiskMachine2018,sablayrollesWhiteboxVsBlackbox2019,Liu23tdsc} focus on the training losses. Additionally, some studies~\cite{choquette-chooLabelOnlyMembershipInference2021,liMembershipLeakageLabelOnly2021} employ the prediction labels for their attacks. These methods are usually considered score-based attack methods.
On the other hand, some studies focus on enhancing attack performance by modeling prediction distributions of the target models~\cite{carlini2022membership,ye2022enhanced}. These methods aim to model the distributions for both member and non-member data. They then leverage the distribution difference to attain superior attack outcomes. These methods are commonly referred to as reference-based attack methods.
such as reference models. 
Other research extends their focus into various scenarios~\cite{liuEncoderMIMembershipInference2021,gaoSimilarityDistributionBased2023,yuanMembershipInferenceAttacks2022,zhang23tdsc} or proposes defense methods against the attacks~\cite{chenRelaxLossDefendingMembership2022,yangPurifierDefendingData2023,huang22tdsc,liu23tdscdefense,hu23tdscdef}.
In the experiments, we consider two representative attack approaches to evaluate model privacy leakage: 
score-based~\cite{salemMLLeaksModelData2019,LWHSZBCFZ22} and reference-based~\cite{carlini2022membership} membership inference attacks.

\noindent \textbf{Enhanced membership inference attacks.} More recently, researchers have begun incorporating additional information as key indicators to boost the overall effectiveness of their attacks.
For instance, in the work by He et al. \cite{heSemiLeakMembershipInference2022}, the adversary leverages prediction outcomes obtained from multiple augmented views to significantly enhance its performance. 
Another study by Li et al. \cite{liAuditingMembershipLeakages2022} focuses on results derived from multi-exit models as their attack strategy.
Furthermore, the study conducted by Hu et al.~\cite{hu2022m} integrates prediction results from a multi-modality model to achieve enhanced attack performance.
Inspired by previous studies, we propose a novel approach to enhance attack performance by leveraging additional information from fairness interventions.
Differently, 
our method uniquely exploits the disparities introduced by model fairness techniques and integrates this insight with existing attack strategies. This innovative combination yields superior attack results, revealing previously unrecognized vulnerabilities in fair models.

\subsection{Attribute inference attacks}
Attribute Inference Attacks (AIAs) aim to infer sensitive attributes of samples using deployed model predictions. These attacks share similarities with MIAs, particularly in how some AIA methods exploit prediction gaps between different subgroups to infer attribute information, as demonstrated by Yeom et al. and Ganju et al.~\cite{yeom2018privacy,ganju2018property}. These approaches are typically classified as black-box attacks.
However, state-of-the-art AIAs often employ more sophisticated techniques, relying on the embeddings of target samples. These white-box attacks infer sample attribute information by analyzing the extracted features from the target model.
In our comprehensive privacy evaluations of fair binary classifiers, we consider both types of attack methods: those exploiting prediction gaps (black box) and those leveraging the embeddings (white box). This dual approach allows us to assess the vulnerabilities of fair models from multiple perspectives, providing a more thorough understanding of their privacy implications.

\subsection{Fairness interventions and attacks}
Currently, limited research focuses on algorithmic fairness and attacks. Some earlier studies have explored the connections between fairness studies and adversarial attacks~\cite{xu2021robust,zhang2023revisiting,Chang2020OnAB,ijcai2023p59,mehrabi2021exacerbating}. They find that fair models tend to be more vulnerable to attacks than biased models~\cite{Chang2020OnAB}. Later, studies have taken the approach to attack target models and compromise model fairness results~\cite{zhang2023revisiting,ijcai2023p59,mehrabi2021exacerbating}.

Recent research has explored various methods to attack fair-enforced models. For instance, studies by Aalmoes et al.~\cite{aalmoes2022leveraging} and Balunović et al.~\cite{balunovic2022fair} have investigated the relationship between attribute attacks and fair-enforced models. Balunović et al.~\cite{balunovic2022fair} aim to promote fairness predictions by reducing the performance of attribute attacks, while Aalmoes et al.~\cite{aalmoes2022leveraging} utilize fairness methods to defend against such attacks. These studies demonstrate the alignment between model fairness and attribute privacy. In contrast, our work uncovers an overlooked attack mechanism that poses significant threats to attribute privacy in fair-enforced models.

More related to our study, Chang and Shokri \cite{chang2021privacy} attack fair-enforced methods with membership inference attack methods. They consider fairness constraint methods for decision tree models with structure data. They then measure the attack performance with average-case success metrics of accuracy and AUC. They find that score-based methods can effectively attack fair models more accurately than biased ones. Our study aims to examine the attack performance in binary classifications and enforce more efficient attacks. We employ multiple attack methods and metrics to assess the privacy impact of fairness approaches.

\section{Preliminaries}
\subsection{Algorithmic fairness}
Given biased models, we consider sensitive attributes $s \in \mathcal{S}$ and subgroups $\{s_0, s_1\}$ with binary attribute values $\{0,1\}$. 
Then, for fair models, as the prediction target and the sensitive attribute are irrelevant, the model prediction $ \widetilde{y}$ and $S$ should be independent, \textit{i.e.}, $\widetilde{y} \perp S | Y=y$. 

With different values of the sensitive attribute $\{s_0, s_1\} \in A$, one selected fairness metric $\gamma$ can be expressed as follows: 
\begin{equation}
    \gamma (\widetilde{y}, y, S), S=\{s_0, s_1\}
\end{equation}

\noindent \textbf{Fairness metrics.} Ideally, the fairness metric value for different subgroups should be equal ($\gamma_{s_0} = \gamma_{s_1}$). 
However, biased models tend to have different metric values across subgroups. 
Generally, we adopt the difference to quantify the discrimination level and measure model fairness performance:
\begin{equation}
    \Gamma = |\gamma_{s_0} - \gamma_{s_1}|,
\label{eq:g}
\end{equation}
where $\Gamma$ is the discrimination level, $\gamma_{s_0}$ and $\gamma_{s_1}$ are the fairness metric values across subgroups. 
In the experiments, we adopt bias amplification (BA) from~\cite{zhao-etal-2017-men} and equalized odds (EO) from~\cite{Hardt:2016wv} as fairness metrics.

\textit{Bias amplification} (BA) is introduced in~\cite{zhao-etal-2017-men}. BA measures the difference in true positive predictions across subgroups, normalized by the total true positives. A lower BA indicates less bias amplification. It can be written as:
\begin{equation}
        \frac{1}{2} \frac{\left|\mathrm{TP}_{s_0}- \mathrm{TP}_{s_1}\right|}{\mathrm{TP}_{s_0}+\mathrm{TP}_{s_1}},
\end{equation}
where $\text{TP}$ presents the true positive value of predictions.

\textit{Equalized odds} (EO) is proposed in~\cite{Hardt:2016wv}. EO requires that the probability of a positive prediction given the true label should be equal across subgroups. This ensures that both true positive rates (TPRs) and false positive rates (FPRs) are equalized across subgroups. The fairness metric can be defined as:
\begin{equation}
        P{\{\widetilde{Y}|Y, S\}}; Y=\{0,1\} , S=\{s_0, s_1\}.
\end{equation}
We report the discrimination level with the difference of BA and EO across subgroups for fairness evaluations as they are widely adopted for fairness evaluations. For instance, with the discrimination level in Eq.~(\ref{eq:g}), fairness measurement considering the metric of EO can be calculated as follows: 
\begin{equation}
    \text{DEO} = |\text{EO}_{s_0} - \text{EO}_{s_1}|.
\end{equation}

\noindent \textbf{Datasets.} 
CelebA~\cite{CelebAMask-HQ}, UTKFace~\cite{geraldsutkface}, and FairFace~\cite{karkkainen2021fairface} are the commonly adopted datasets in fairness studies. CelebA contains over $200,000$ celebrity face images with $40$ attribute annotations. UTKFace and FairFace are diverse facial datasets with balanced annotations across different demographic groups. In our experiments, we create biased training data by sampling with a 9:1 ratio between majority and minority groups while maintaining balanced distributions for the target classification tasks and test sets. In the evaluations, we first adopt the CelebA dataset in Sections 4 and 5, focusing on smiling classifications as the target and gender as the sensitive attribute. We then extend our analysis in Section 7 to include all three datasets, using various attribute combinations to further validate the generalizability of our observations and the efficacy of our proposed attack methods across different data distributions and fairness scenarios.

\subsection{Membership inference attacks}
In the evaluations, we consider score-based and reference-based MIA methods to evaluate model privacy.

\noindent \textbf{Score-based attack methods} rely on the target model’s (i.e., models under attack) prediction outcomes (e.g., scores or losses) to determine the membership on each individual data sample.
Typically, to mimic the behavior of the target model, a ``shadow model" is trained with an auxiliary dataset that shares the same distribution as the training data.
The outputs of the shadow model are then adopted to train the attack models, 
where the membership of the data is considered as the labels.
In this way, 
the attack model can infer whether the given samples are from the training data or not.
Formally, 
given target models $\mathcal{T}$ with queried sample $x$, 
the membership of the sample $M(x)$ can be predicted by,
\begin{equation}
    M(x) = f_a^m (\mathcal{T}(x)),
\label{mia_s}
\end{equation}
where the designed \textit{membership} attack model $f_a^m$ outputs the confidence scores of predicted membership. 
Generally, 
existing studies usually adopt deep learning models as attack models.

\noindent \textbf{Reference-based likelihood ratio attack methods}, 
on the other hand, 
infer the membership by modeling the prediction distributions. 
They first train multiple shadow models on random subsets of training data. 
For a target example $x$, 
the methods then model the prediction distributions for models ($f_{\text{in}}$) trained with the sample $x$ and models ($f_{\text{out}}$) trained without $x$. 
Both distributions are modeled as Gaussians. 
Then, 
they determine the membership of $x$ by comparing the likelihood of the sample prediction results $\mathcal{T}(x)$ from the target model with the two distributions above. 
Formally, 
the likelihood ratio between the distributions of member and non-member data can be defined as,
\begin{equation}
   \Lambda=\frac{p(\phi(\mathcal{T}(x))|\mathcal{N}(\mu_{\text{in}}, \sigma_{\text{in}}))}{p(\phi(\mathcal{T}(x))|\mathcal{N}(\mu_{\text{out}}, \sigma_{\text{out}}))},
\label{lira}
\end{equation}
where $\phi$ is a logic scaling function, 
$(\mu_{\text{in}}, \sigma_{\text{in}})$ are calculated with the predictions from the predictions of member data ($f_{\text{in}}$), and  $(\mu_{\text{out}}, \sigma_{\text{out}})$ are from $f_{\text{out}}$. 
With likelihood ratio $\Lambda$, 
whichever is more likely determines the membership of $x$.

\subsection{Attribute inference attacks}
AIAs aim to infer sample sensitive information--subgroups such as male or female. They can be conducted in both black-box and white-box settings. In a black-box setting, given predictions of target models, the subgroup can be predicted by:
\begin{equation}
    A(x) = f_a^a (\mathcal{T}(x)).
\label{aia_s}
\end{equation}
Similar to MIAs, the designed \textit{inference} attack model $f_a^a$ predicts sample subgroups with target model predictions. 

As state-of-the-art AIAs usually rely on sample embedding of the target
models with a white-box attack setting. Specifically, the trained target model $\mathcal{T}$ can be composed of a feature extraction module $h$ and a classification module $g$, such that $\mathcal{T}=g(h(x))$. Given sample embedding $h(x)$, the subgroup can be predicted by:
\begin{equation}
    A(x) = f_a^a (h(x)).
\label{aia_f}
\end{equation}

\subsection{Notations}
Table~\ref{tab:notation} summarizes the main notations adopted throughout the paper.
In the table, $x$ and $y$ presents model inputs and labels, respectively. Predicted labels are represented as $\widetilde{y}$. Our models consist of feature extractors $h(x)$, and classification heads $g(\cdot)$.
In our privacy analysis, membership inference results for a sample $x$ are denoted as $M(x)$, while attribute inference results are represented as $A(x)$. The attack models for membership and attribute inference are denoted as $f_a^m(\cdot)$ and $f_a^a(\cdot)$, respectively. We consider different types of attacks: score-based ($\text{MIA}_\text{s}$) and reference-based ($\text{MIA}_\text{l}$) membership inference attacks, as well as black-box ($\text{AIA}_\text{b}$) and white-box ($\text{AIA}_\text{w}$) attribute inference attacks.
We adopt metrics including accuracy of target classifiers ($\text{Acc}_t$) and attack classifiers ($\text{Acc}_a$), as well as true positive rate ($\text{TPR}$) and false positive rate ($\text{FPR}$).

\begin{table*}[t]
\centering
\caption{Summary of Main Symbols and Notation}
\label{tab:notation}
\begin{tabular}{ll ll}
\toprule
\textbf{Symbol} & \textbf{Description} & \textbf{Symbol} & \textbf{Description} \\
\cmidrule(lr){1-2}\cmidrule(lr){3-4} 

$x$ & Inputs & $y$ & Labels \\

$\widetilde{y}$ 
& Predicted labels
& $h(x)$ 
& Feature extractors
\\

$g(\cdot)$ 
& Classification heads
& $\mathcal{T} = g \circ h$ 
& Target models ($\mathcal{T}(x) = g(h(x))$) 
\\

$s \in \mathcal{S}$ 
& Sensitive attributes 
& $s_0, s_1$
& Binary attribute values (0, 1)
\\

$\gamma(\widetilde{y}, y, S)$ 
& Generic fairness metric 
& $\Gamma = |\gamma_{s_0} - \gamma_{s_1}|$ 
& Discrimination level across subgroups
\\

$\mathrm{BA}$ 
& Bias Amplification (fairness metric) 
& $\mathrm{DEO}$ 
& Difference in Equalized Odds (fairness metric)
\\

$M(x)$ 
& Membership inference results for $x$ 
& $A(x)$ 
& Attribute inference results for $x$ 
\\

$f_a^m(\cdot)$ 
& Membership attack models
& $f_a^a(\cdot)$ 
& Attribute attack models
\\

$\text{MIA}_\text{s}$ 
& Score-based membership inference attacks
& $\text{MIA}_\text{l}$ 
& Reference-based membership inference attacks (LiRA)
\\

$\text{AIA}_\text{b}$ 
& Black-box attribute attacks  
& $\text{AIA}_\text{w}$ 
& White-box attribute attacks
\\

$\phi(\cdot)$ 
& Logit scaling functions
& $p(\cdot)$ 
& Probability functions
\\

$\mu_{\text{in}}, \mu_{\text{out}}$ 
& Mean prediction vectors (member, non-member) 
& $\sigma_{\text{in}}, \sigma_{\text{out}}$ 
& Std.\ dev.\ (member, non-member)
\\

$\text{Acc}_t$ 
& Accuracy of target (main) classifiers 
& $\text{Acc}_a$ 
& Accuracy of attack classifiers
\\

$\text{TPR}$ 
& True positive rate 
& $\text{FPR}$ 
& False positive rate
\\
$\Lambda$ 
& Likelihood ratio in reference-based MIA 
& $\mathrm{Cov}$ 
& Covariance matrix for distribution modeling
\\

$\mathcal{T}_b(x), \mathcal{T}_f(x)$ 
& Biased, fair target models 
& $h_b(x), h_f(x)$ 
& Biased, fair feature extractors
\\

$\mathrm{TP}_{s_0}, \mathrm{TP}_{s_1}$ 
& True Positives for subgroups $s_0, s_1$ 
& $\mathrm{AUC}$ 
& Area-under-ROC-curve measure
\\

\bottomrule
\end{tabular}
\end{table*}

\section{Evaluations with MIAs}
In this section, we evaluate the privacy impact of fairness interventions with MIAs. We first introduce attack settings and then present the attack results.  

\subsection{Attack settings}
\begin{figure}[t]
\begin{center}
\includegraphics[width=.9\linewidth]{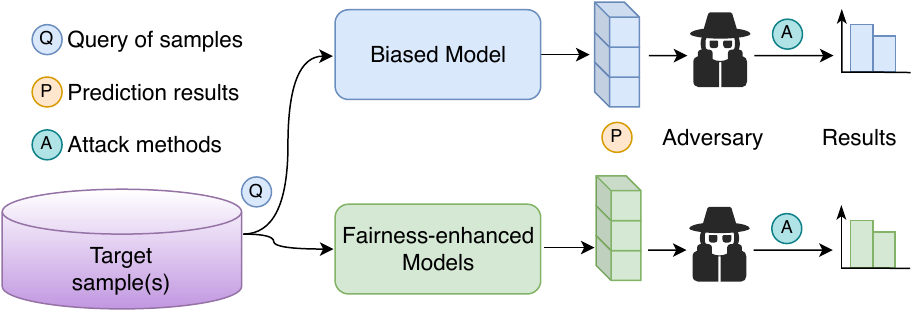}
\end{center}
\caption{Model privacy impact evaluation pipelines.}
\label{fig:pipeline}
\end{figure}

\noindent \textbf{Attack pipeline.} 
We evaluate the privacy performance of models before and after applying fairness interventions using Membership Inference Attacks (MIAs). We follow the attack pipeline depicted in Figure~\ref{fig:pipeline}, which is consistent with the approach adopted in previous work~\cite{chang2021privacy}.
Specifically, we begin by training biased models using biased training data. To enhance the fairness performance of these models, we apply fairness interventions to obtain their counterpart, fair models.
Next, we attack both the biased and fair models using existing MIA methods.
Finally, we compare the attack results to evaluate the impact of fairness interventions on the privacy of the models.

\noindent \textbf{Target models.} We train biased models with the CelebA dataset \cite{CelebAMask-HQ}, which contains imbalanced data distributions for various attributes. In particular, we consider \textit{smile} as classification targets and \textit{gender} as the sensitive attribute. 
We train biased models following settings in \textit{ML-Doctor} from~\cite{LWHSZBCFZ22}.
We apply fair mixup operations from~\cite{mroueh2021fair,du2021fairness} to mitigate the biased predictions. 
Table \ref{table:score_results} presents accuracy ($\text{Acc}_\text{t}$) and fairness metrics ($\text{BA}$, $\text{DEO}$)
results for both biased (``Bias'') and fair (``Fair'') models.
The results show decreased fairness metric results, indicating the effectiveness of the adopted fairness interventions.

\noindent \textbf{Threat models.}
We apply both the score-based attacks (MIAs from~\cite{LWHSZBCFZ22}) and reference-based attacks (LiRA from~\cite{carlini2022membership}) on target models in a black-box manner. In particular, adversaries can only access models' predictions and an auxiliary dataset, which shares similar data distributions with the training data.
The adversary trains shadow models to mimic the target models' behavior and uses the prediction scores and results (\textit{true or false predictions}) to infer sample membership. 
We conduct the attacks following settings in \textit{ML-Doctor}.

\subsection{Score-based attacks}
Table~\ref{table:score_results} shows the $\text{Acc}_\text{a}$ and $\text{AUC}_\text{a}$ results for attacks on the models. It shows improved attack results after fairness interventions. For example, the accuracy results decreased from $59.8\%$ to $53.2\%$ with the fair models. AUC results exhibit similar trends. This aligns with results in Figure~\ref{fig:intro}, where fewer training samples can be successfully attacked after the interventions. Our results show that fairness interventions provide some defense against existing MIAs. 

\begin{table}[t]
\caption{Attack results with the score-based methods from~\cite{LWHSZBCFZ22} in (\%).}
\centering 
  \begin{tabular}{l r r r r r}
  \toprule{Models}&
   {$\text{Acc}_\text{t}$ $\uparrow$} &
   {$\text{BA}$ $\downarrow$}&
   {$\text{DEO}$ $\downarrow$}&
   {$\text{Acc}_\text{a}$ $\uparrow$} &
   {$\text{AUC}_\text{a}$ $\uparrow$} 
    \\
    \midrule{Bias} &87.6 &7.7  &21.7  &59.8 &62.8  \\
            {Fair} &90.5 &2.5 &5.6 &53.2 &54.8  \\
    \bottomrule
  \end{tabular}
  \label{table:score_results}
\end{table}

\subsection{Reference-based attacks}
We further use the reference-based attack method of LiRA from~\cite{carlini2022membership} to evaluate model privacy. The method examines attack performance via the True Positive Rates (TPR) value in the low False Positive Rates (FPR) region. This enables MIAs on hard examples, where samples from both member and non-member groups share similar prediction results. Table~\ref{table:lira_results_case} shows the attack results with the TPR results at a low FPR value of $0.1\%$. The table shows inferior attack performance for fair models compared to the biased ones with all three considered metrics. The results are consistent with the score-based MIAs. 

\begin{table}[t]
\caption{Attacks using LiRA from~\cite{carlini2022membership} with TPR @ 0.1\% FPR  in (\%).}
\centering 
  \begin{tabular}{ c c c c}
  \toprule {Models}&
   {$\text{Acc}_\text{a}$ $\uparrow$} &
   {$\text{AUC}_\text{a}$ $\uparrow$} &
   {TPR $\uparrow$} 
   \\
   \midrule
   {$\text{Bias}$} &51.5  &51.4 &0.6  \\
   {$\text{Fair}$} &50.8  &50.3 &0.2 \\
   \bottomrule
  \end{tabular}
  \label{table:lira_results_case}
\end{table}

\subsection{Discussions}
\begin{figure}[t]
    \centering
    \subfloat[Attack accuracy \label{fig:flipper}]{
        \includegraphics[width=0.23\textwidth]{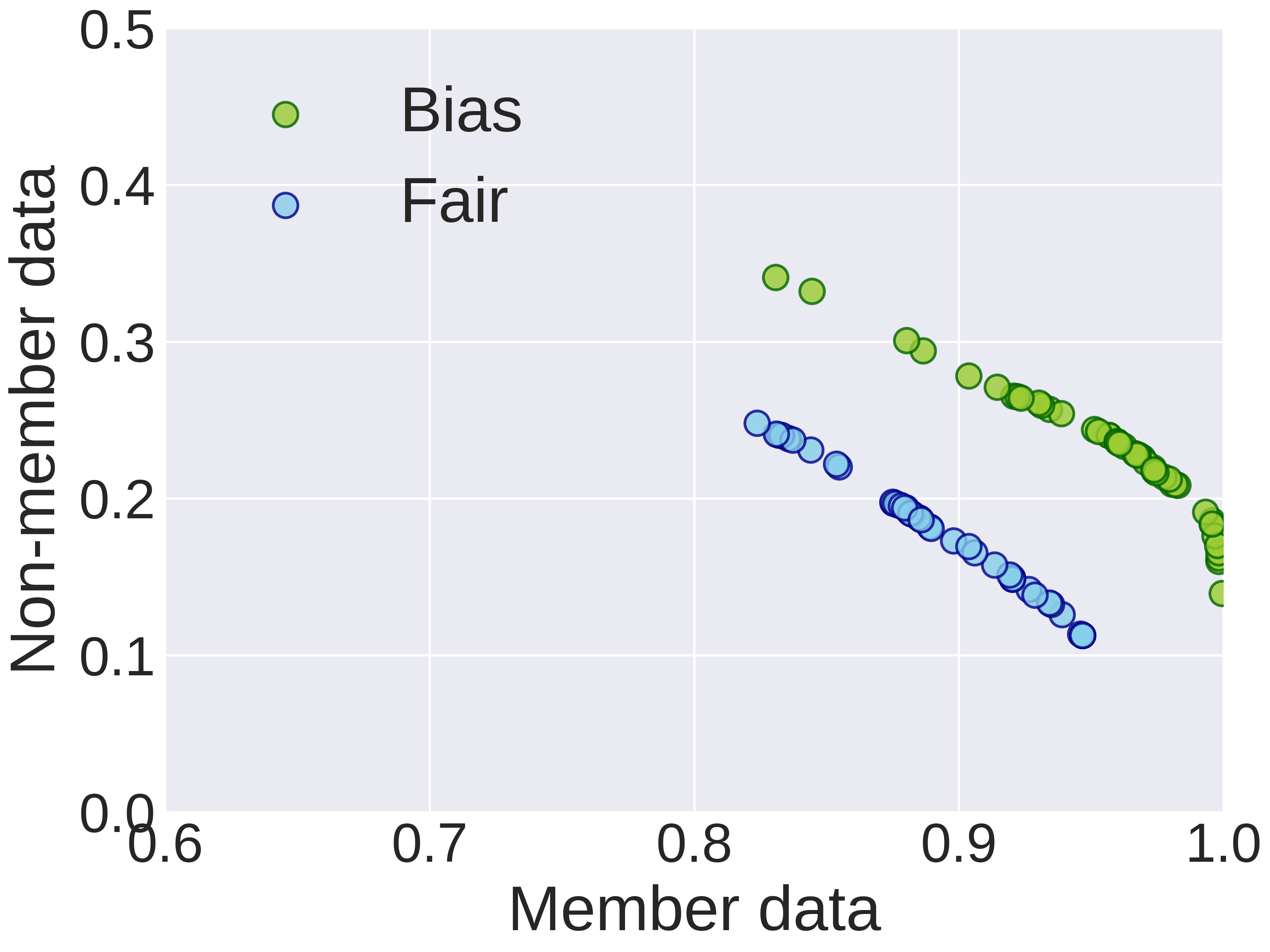}
    }
    \subfloat[Attack for hard examples \label{fig:score_fpr}]{
        \includegraphics[width=0.23\textwidth]{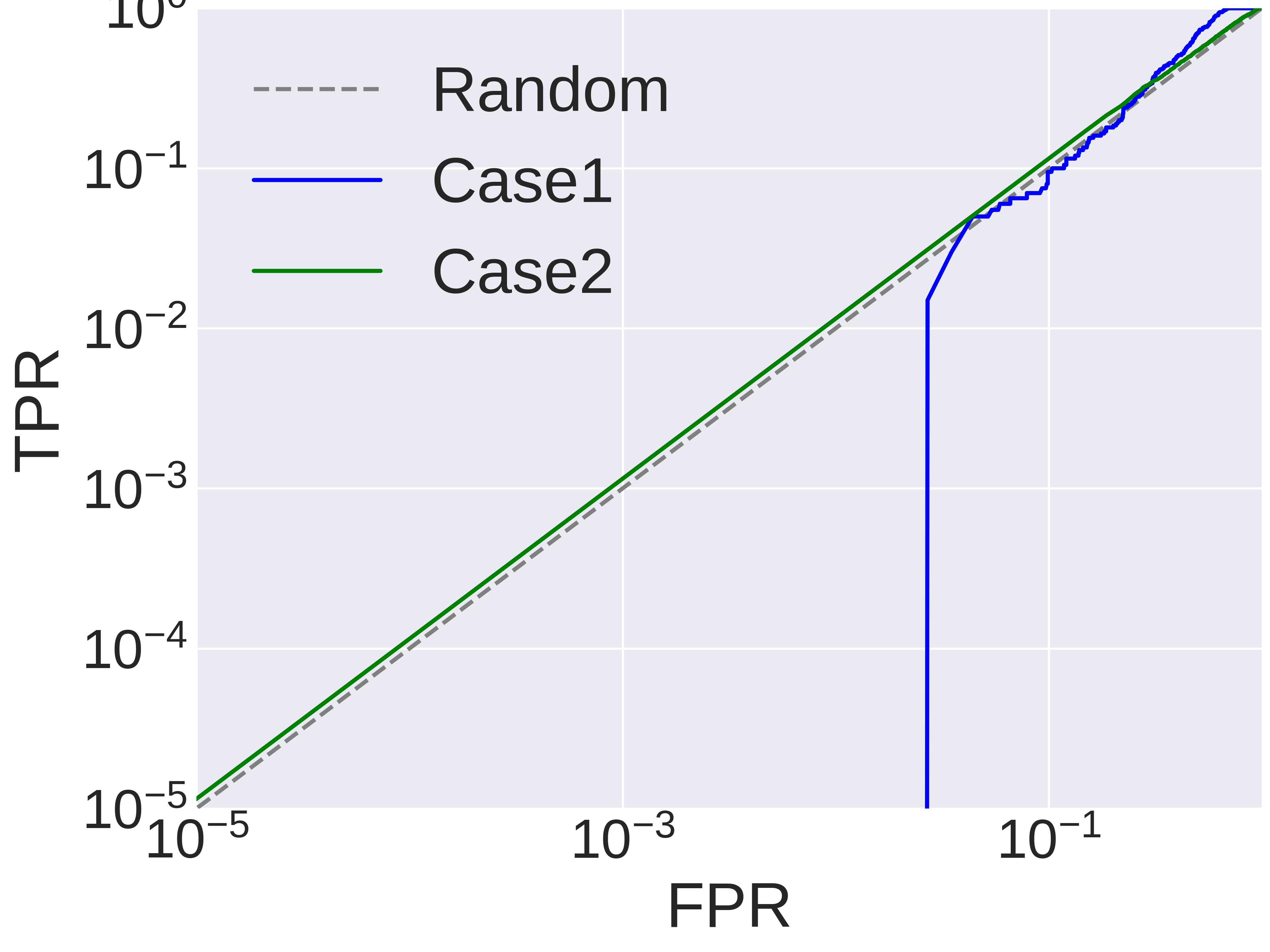}
    }
    \caption{ Existing attacks (a) exhibit clear performance trade-offs between member and non-member data, each green circle represents attack accuracy on biased models and each blue circle represents attack accuracy on fair models; and (b) are inefficient in attacking hard examples in the low FPR region.}
    \label{fig:trade_off}
\end{figure}

To better understand the results, we further explore the attack results and find the following observations:

\noindent \textbf{Performance trade-offs.} 
During the evaluation, 
we observe
\textbf{\textit{evident trade-offs in attack performance on member versus non-member data}}. 
Figure~\ref{fig:flipper} illustrates the inherent performance trade-offs in membership inference attacks by plotting the accuracy results for member data (x-axis) against the accuracy for non-member data (y-axis). We conducted over $100$ independent attack experiments on both biased and fair models, with each circle in the scatter plot representing a single attack instance. Green circles represent attack results on biased models, while blue circles represent attack results on fair models.
The clear negative correlation visible in both model types demonstrates a fundamental trade-off: as an attack model becomes more accurate at identifying member groups (moving rightward on the x-axis), it simultaneously becomes less accurate at correctly classifying non-members (moving downward on the y-axis). This pattern holds consistently across both biased and fair models.
This trade-off raises significant concerns about the practical effectiveness of membership inference attacks. Specifically, it suggests that achieving high attack performance on training data members inevitably comes at the cost of a higher false positive rate (FPR) on non-member data, making the attack less reliable overall. This observation aligns with findings from previous studies~\cite{carlini2022membership} that highlight the limitations of threshold-based attack methods when applied to binary classification tasks.

The issue becomes more pronounced for hard examples where members and non-members share similar prediction scores. 
As suggested in~\cite{carlini2022membership}, we assess the attack performance for hard examples with TPR values in the low FPR region.
We find the TPR values are around $0.0$ for most attacks. 
Figure~\ref{fig:score_fpr} presents two worst-case scenarios. The green curve in the figure shows closely aligned TPR and FPR values, indicating the attack results are equivalent to random guesses.
The blue line shows $0.0$ TPR values in low FPR regions, indicating that no positive samples can be correctly identified.
The findings reveal that attack models fail to differentiate the membership of hard examples,
indicating invalid attacks.
This aligns with the concerns about the effectiveness of score-based attacks raised in previous studies~\cite{carlini2022membership,ye2022enhanced}. 

\begin{figure*}[t]
\begin{center}
\centering
    \subfloat[Biased models]{
        \label{fig:score_bias}
        \includegraphics[width=0.24\textwidth]{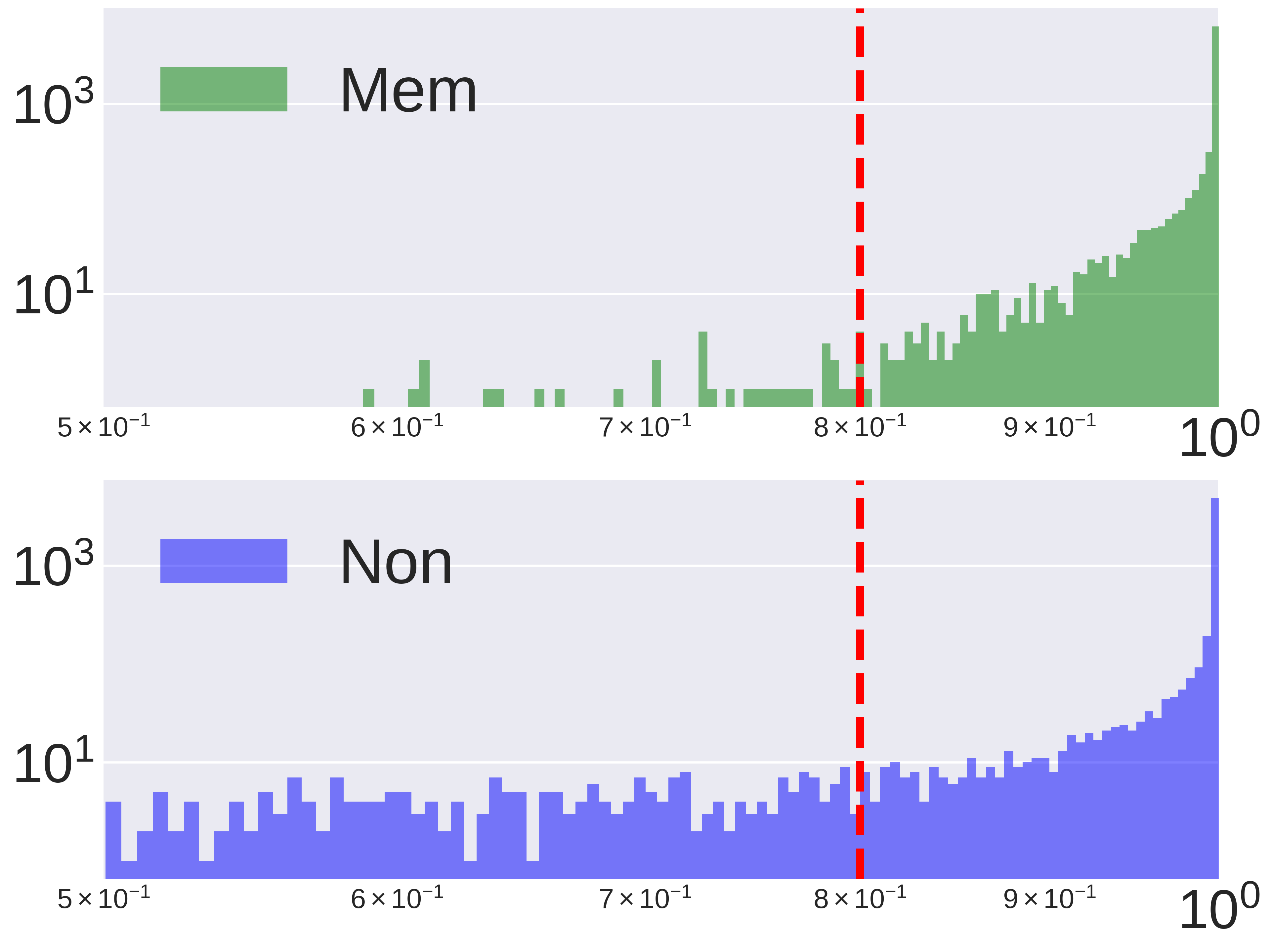}
    }
    \subfloat[Fair models]{\label{fig:score_fair}
        \includegraphics[width=0.24\textwidth]{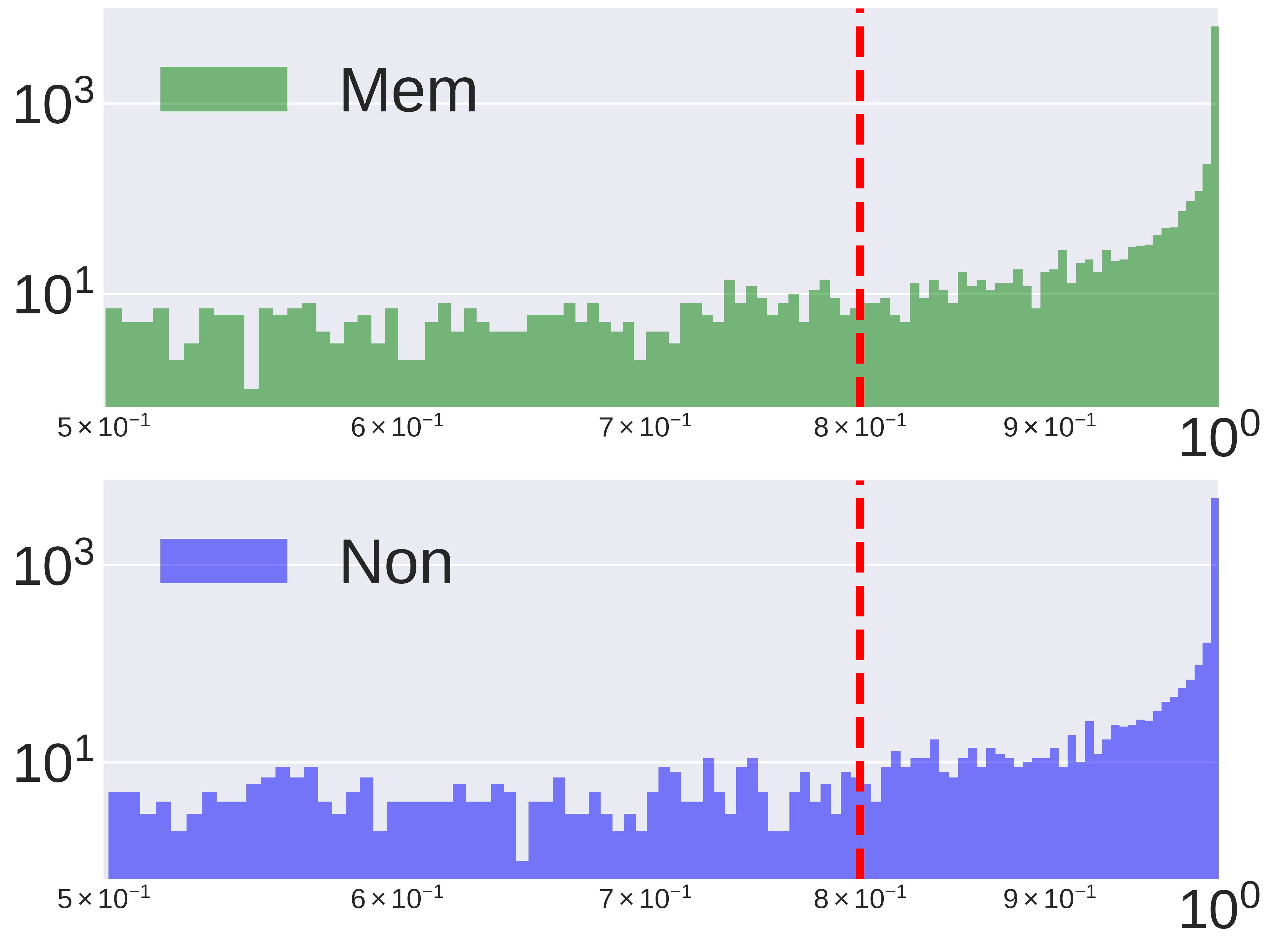}
    }
    \subfloat[Majority subgroups]{\label{fig:subg_majority}
        \includegraphics[width=0.24\textwidth]{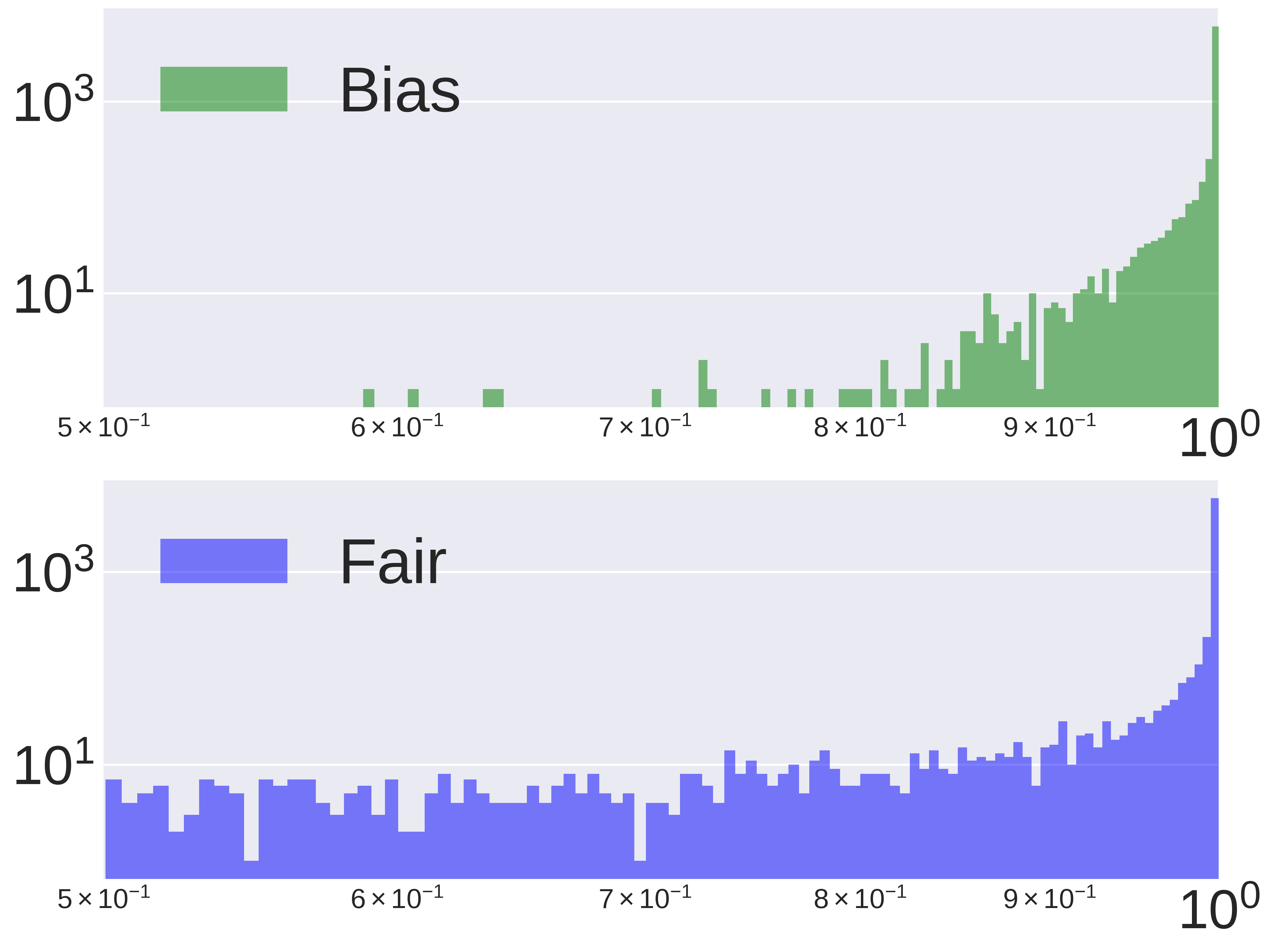}
    }
    \subfloat[Minority subgroups]{\label{fig:subg_minority}
        \includegraphics[width=0.24\textwidth]{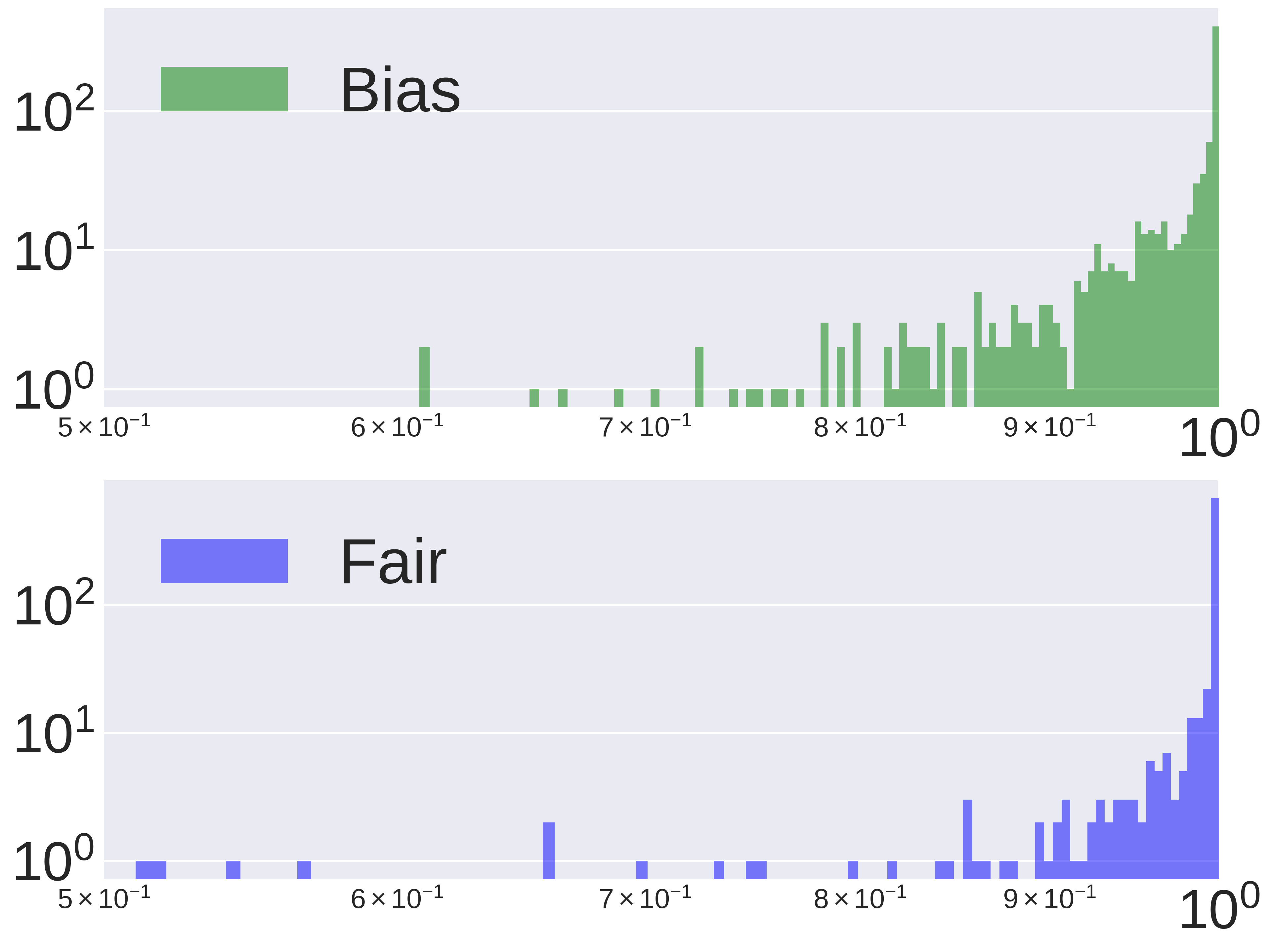}
    }
\end{center}
\caption{Prediction score changes after applying fairness methods. The \textit{red lines} in (a) and (b) indicate that the trained attack models infer sample membership with certain threshold values. (c) and (d) show the changes in terms of different subgroups.}
\label{fig:conf_score}
\end{figure*}

\noindent \textbf{Model degradation.} 
To explore the reason for the trade-off phenomenon, we have discovered that \textbf{\textit{trained attack models typically degrade into simple threshold models with one-dimensional inputs.}} 
This is because current attack methods rely on prediction outcomes to determine the sample membership. 
For binary classifiers, 
prediction scores can be reduced to one dimension as the sum of the confidence scores always equals one. 
Consequently, 
the attack model can essentially be viewed as a simple threshold model, 
which infers the membership by ``thresholding'' one-dimensional values. 

Figure~\ref{fig:score_bias} presents histograms of prediction scores with vertical lines indicating the threshold value. 
By adjusting the vertical line (thresholds), 
it is possible to achieve higher accuracy for member data, 
but this comes at the expense of decreased accuracy for non-member data. 
This threshold adjustment explains the trade-off phenomenon.

\noindent \textbf{Impacts of fairness interventions.} 
When examining the prediction scores, 
we find that \textbf{\textit{fairness interventions decrease confidence scores for the majority training data, introducing some defense against existing MIAs.}}
This is evidenced by the histograms of confidence scores in Figures~\ref{fig:score_bias} and~\ref{fig:score_fair}.
The figures show that fairness interventions result in more similar score distributions between member and non-member data, making it more difficult for the threshold-based attack models to distinguish them.

Moreover, 
we explore the score changes for different subgroups in Figures~\ref{fig:subg_majority} and~\ref{fig:subg_minority}. 
From the plots, the majority data are more ``spread out'', whereas the minority are more ``concentrated''.  
This is because fairness interventions strive to balance prediction performance across subgroups for fair predictions.
The results advocate the observed increased loss values for most data points in Figure~\ref{fig:intro_mia}.
It also aligns with the fairness-utility trade-off, which is extensively observed in fairness studies~\cite{zhang2023fairness,PinzonPPV22,Zietlow_2022_CVPR}.

Our analyses indicate that existing attack methods are ineffective in exploiting prediction gaps that could lead to model privacy leaks. While fairness interventions do introduce some defense to MIAs, we identify a novel threat that will pose significant risks to model privacy.

\section{Evaluations with AIAs}
This section further assesses model privacy performance by conducting Attribute Inference Attacks (AIAs). We begin by outlining the attack settings, followed by a presentation of the attack results and an in-depth analysis of our findings. Our evaluation uses the same CelebA dataset configuration as in the previous section to ensure consistency. 

\subsection{Attack settings}
\noindent \textbf{Attack pipeline.} Attribute Inference Attacks (AIAs) are designed to infer sensitive attribute information about data samples by exploiting the outputs of target models. We follow the same attack pipeline in Figure~\ref{fig:pipeline}. 

\noindent \textbf{Target models.} We consider the same target models as in MIAs. Specifically, we launch AIAs on both biased and fair models and compare the attack results for privacy evaluations.

\noindent \textbf{Threat models.}
In our evaluations, we employ both black-box and white-box AIAs to comprehensively assess the privacy risks of biased and fair models.

In the \textit{black-box} setting, the attacker has access only to the prediction scores from the target model. This scenario mimics real-world situations where an adversary can query the model but has no access to its internal structure. In the \textit{white-box} setting, the attacker has more privileged access and can obtain sample embeddings from internal layers of the target model. Specifically, we extract features from the \textit{last} layer of the feature extraction module in the target model, which is the layer immediately preceding the fully connected layers. This approach provides the attacker with more information with the sample embeddings. For both settings, we train attack models following the methodology outlined in the \textit{ML-Doctor} framework \cite{LWHSZBCFZ22}. 


\subsection{Attack results}
Table~\ref{table:aia_results} presents the attack results, where $\text{AIA}_\text{b}$ represents the black-box attack, and $\text{AIA}_\text{w}$ represents the white-box attack. For example, the black-box attack accuracy decreases from $56.5\%$ for biased models to $47.5\%$ for fair models, and the white-box attack accuracy decreases from $83.4\%$ for biased models to $82.9\%$ for fair models. Notably, the black-box attack accuracy on fair models is close to random guessing. 

\begin{table}[t]
\caption{Attribute inference attack results with biased and fair models in (\%).}
\centering 
  \begin{tabular}{l r r r r r}
  \toprule{Models}&
   {$\text{Acc}_\text{t}$ $\uparrow$} &
   {$\text{BA}$ $\downarrow$}&
   {$\text{DEO}$ $\downarrow$}&
   {$\text{AIA}_\text{b}$ (Acc) $\uparrow$} &
   {$\text{AIA}_\text{w}$ (Acc) $\uparrow$} 
    \\
    \midrule{Bias} &87.6 &7.7  &21.7  &56.5 &83.4  \\
            {Fair} &90.5 &2.5 &5.6 &47.5 &82.9  \\
    \bottomrule
  \end{tabular}
  \label{table:aia_results}
\end{table}

\subsection{Discussions}
Our results suggest that fairness interventions introduce some level of robustness against AIAs. For black-box attacks, the attack model uses the prediction scores of target models to infer sample subgroup information. As observed in Figure~\ref{fig:subg_majority} and Figure~\ref{fig:subg_minority}, fairness interventions adjust scores across subgroups to achieve fairer predictions. This adjustment results in more similar scores across subgroups, making it harder for attackers to infer subgroup information. Consequently, the pursuit of fairer results can inadvertently enhance model privacy. These findings align with the study by \cite{balunovic2022fair}, which examined model fairness by evaluating whether extracted features contain sensitive information via AIAs.

Similar trends can also be observed with white-box attacks. Fair models resulted in lower attack success rates compared to biased models. Notably, the attack accuracy for white-box attacks remains higher than for black-box attacks, even though it declined after fairness interventions.
This is because white-box attacks leverage sample embeddings to infer subgroup information, preventing trained attack models from degrading into simple threshold models.
As black-box attacks adopt prediction scores as inputs, similar to MIAs, attack models degrade into simple threshold models when considering binary classifiers as the target models. However, with white-box attacks, the inputs are extracted features, ensuring sufficient information to launch efficient AIAs.

\section{An enhanced attack mechanism}
While our experiment results show inferior attack results after fairness interventions, during the evaluations, we discover an overlooked attack mechanism and propose two novel attack methods, FD-MIA and FD-AIA. We first introduce the attack mechanism and then present the proposed FD-MIA and FD-AIA.

\subsection{Fairness disparity based attack mechanism}
\begin{figure}[t]
    \centering
    \subfloat[All data\label{fig:intro_general}]{
        \includegraphics[width=0.24\textwidth]{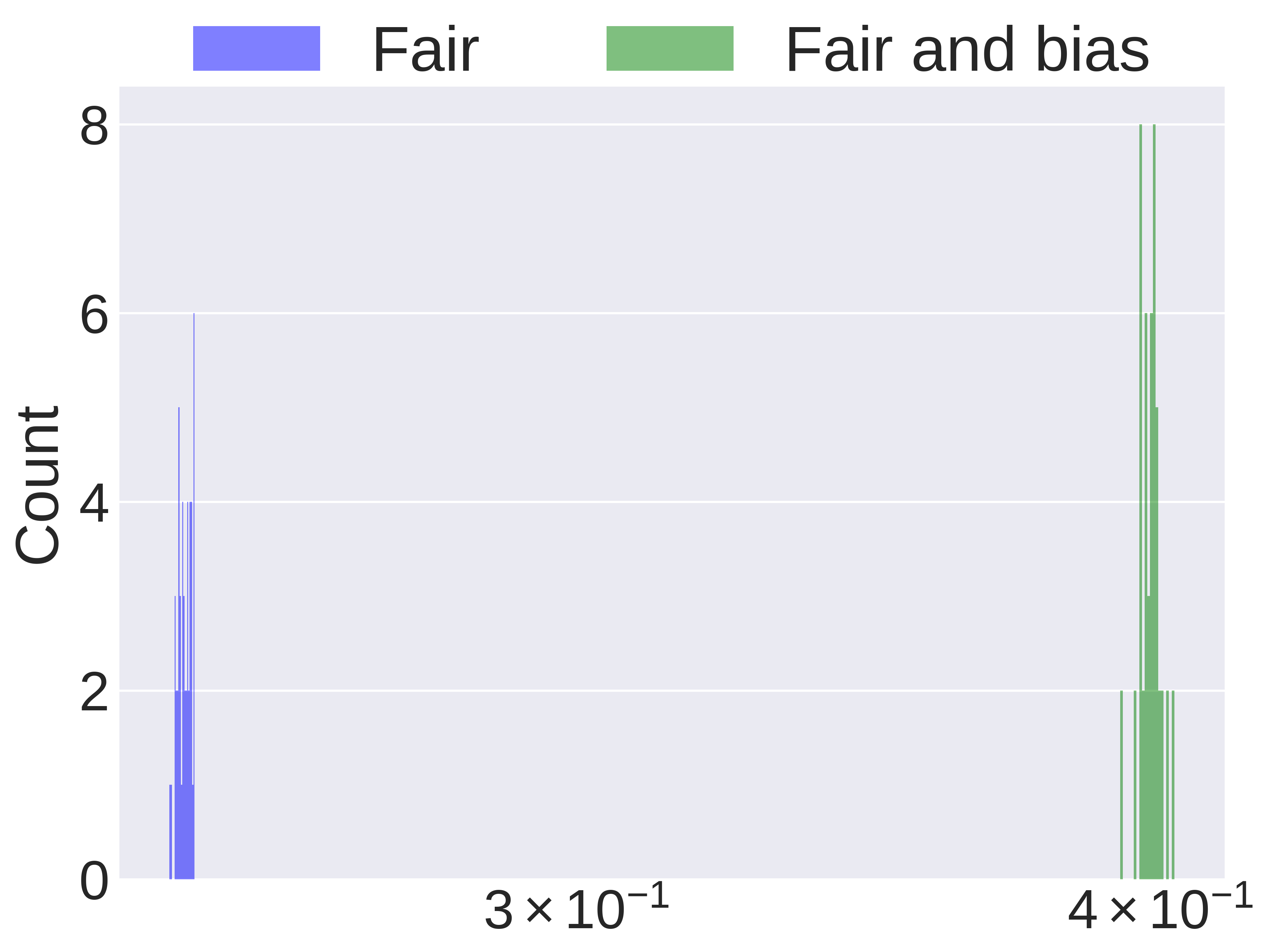}
    }
    \subfloat[Hard examples\label{fig:intro_hard}]{
        \includegraphics[width=0.24\textwidth]{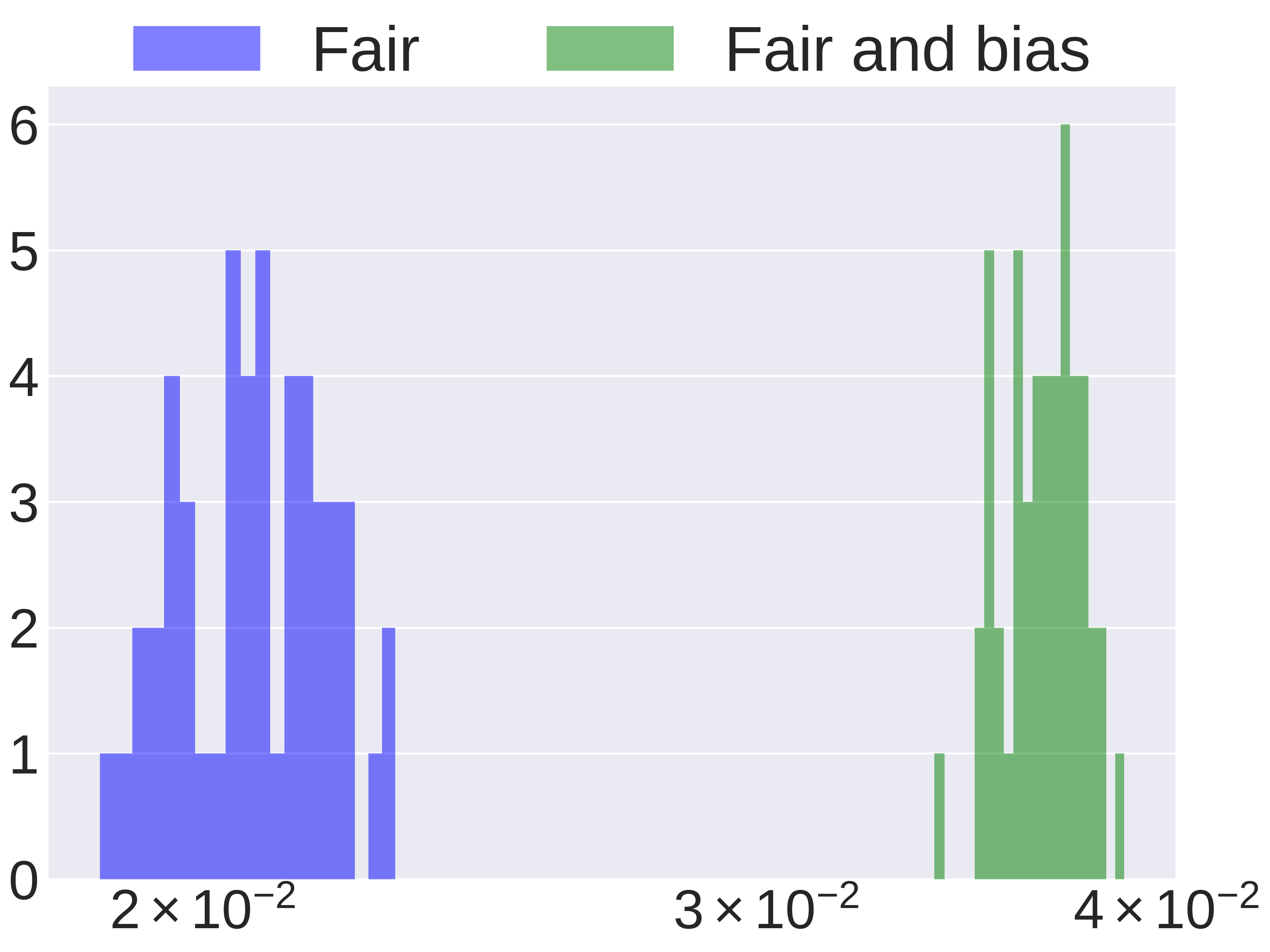}
    }
    \caption{Histograms of prediction score distances between groups of member and non-member data for fair and biased models. 
    We measure the distance with score value difference between the groups and present comparisons regarding (a) all data and (b) hard examples, where samples from the member and non-member data share similar scores.}
    \label{fig:gap}
\end{figure}

\noindent \textbf{The enlarged distribution gaps.}
The previous findings have indicated that fairness methods tend to decrease the score values for the majority subgroups while increasing the scores for the minority subgroups. This can lead to an enlarged gap in predictions across subgroups.
On the other hand, the score changes for the non-member data are likely to follow normal distributions, causing a different behavior pattern compared to the member data.
These different behavior patterns in score changes can serve as additional clues to achieve better performance in MIAs and AIAs.
For example, in MIAs, we can plot the histograms of the prediction gaps, as shown in Figure \ref{fig:gap}, to analyze the gaps for the overall training data and the hard examples.

Specifically, we first calculate the mean and variance of the prediction scores for biased and fair models. We then compute the distribution distance between member and non-member data using the results from the biased models and the results from both biased and fair models, respectively. This calculation is performed over a total of $50$ runs. The figure shows enlarged distances when predictions from both models are considered. Moreover, we explore the distance in Figure~\ref{fig:intro_hard} considering only the hard examples--samples with similar prediction scores among member and non-member data. Again, the figure demonstrates enlarged prediction distances when using predictions from both models.
Inspired by the observations, we propose an enhanced attack method tailored for fair models with the observed prediction gaps.

\begin{figure}[t]
    \centering
    \subfloat[Attack pipeline]{
        \includegraphics[width=.45\textwidth]{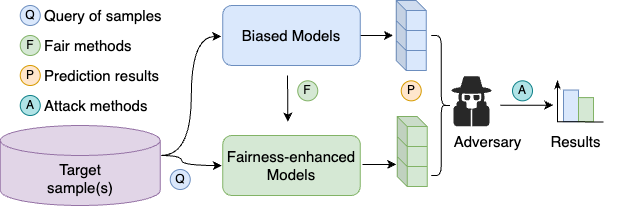}
    } \\
    \subfloat[Enlarged distribution gaps]{
        \includegraphics[width=.5\textwidth]{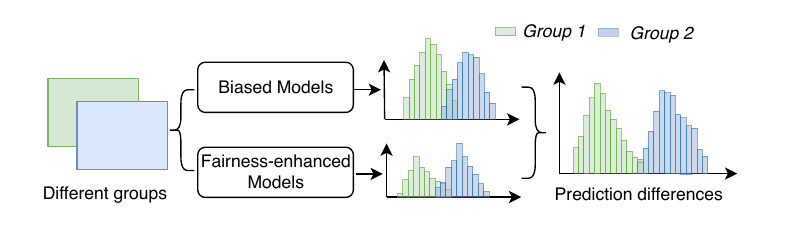}
    }
    \caption{Fairness discrepancy based attacks exploit the difference in predictions from both models to more effectively distinguish different data groups, such as different subgroups or member versus non-member data.}
    \label{fig:dual_flow}
\end{figure}

\noindent \textbf{Attack pipeline.}
Figure~\ref{fig:dual_flow} illustrates the attack pipeline, 
wherein an adversary can access prediction results from both models. 
The attack models will exploit the difference in predictions to infer the membership or subgroup information. 
We refer to the proposed method as the \textit{Fairness Discrepancy based Membership Inference Attack} (\textit{FD-MIA}) and \textit{Fairness Discrepancy based Attribute Inference Attack} (\textit{FD-AIA}). 
As the proposed method only modifies the inputs, 
it can be integrated into existing attack techniques.

\noindent \textbf{Threat models.} The discovered attack mechanism operates as a black-box attack, requiring only access to the predictions from both a biased model and a fair model. In practice, adversaries could obtain such predictions, as real-world models often exhibit persistent biased predictions that linger even after attempts at debiasing. For instance, an attacker could monitor a Machine Learning as a Service (MLaaS) platform over time, as debiasing efforts are typically an ongoing process to adhere to relevant legislation. Alternatively, the adversary could deliberately report biases, compelling the model owner to refine the model in accordance with regulations.
By recording the prediction shifts that occur during these debiasing efforts, the adversary can gather the necessary information to enable efficient attacks that exploit the discrepancies between the biased and fair models.

\subsection{FD-MIA}
The discovered attack mechanism can be seamlessly integrated into the existing MIAs of score-based and reference-based attacks, enhancing their attack performance.

\noindent \textbf{Score-based FD-MIA.} Score-based FD-MIA has been introduced to enhance traditional score-based MIAs by integrating additional encoding layers. These layers are designed to extract the features of model predictions, exploiting the observed prediction gaps. Formally, it can be expressed as follows:
\begin{equation}
    M(x) = f_a^m(\mathcal{T}_{\text{b}}(x),\mathcal{T}_{\text{f}}(x)),
\end{equation}
where the attack models $f_a^m$ takes predictions from both biased models $\mathcal{T}_{\text{b}}$ and fair models $\mathcal{T}_{\text{f}}$.

\noindent \textbf{Reference-based FD-MIA.} Reference-based FD-MIA is integrated with the LiRA framework~\cite{carlini2022membership}, which infer sample membership by modeling the prediction distributions. It enhances attack performance using two target models - the biased and the fair ones. Formally, for a given sample $x$ and target models $\mathcal{T}$, the probability of membership is given by:
\begin{equation}
    p=(\phi(\mathcal{T}(x))|\mathcal{N}(\mu_{\text{b}}, \mu_{\text{f}}, \text{Cov})),
\end{equation}
where $\text{Cov}$ is the covariance matrix. The distribution function $\mathcal{N}$ takes the mean confidence scores from both the biased $\mu_{\text{b}}$ and fair models $\mu_{\text{f}}$. This function estimates the likelihood of a data point being a member or non-member. The result is determined by the higher probability score.

\subsection{FD-AIA}
The designed fairness disparity based attack mechanism can be integrated into existing black-box and white-box attribute inference attacks.

\noindent \textbf{Black-box FD-AIA.} Black-box AIAs infer sample subgroup information with target model prediction results. With results from both biased and fair models, the subgroup information can be predicted as:
\begin{equation}
    A(x) = f_a^a (\mathcal{T}_\text{b}(x), \mathcal{T}_\text{f}(x)).
\end{equation}
Here, the attack model $f_a^a$ combines these prediction results to infer the sensitive subgroup information for the input sample $x$.

\noindent \textbf{White-box FD-AIA.}
On the other hand, white-box AIAs use sample embeddings to obtain subgroup information predictions. With access to both the biased and fair models, the prediction can be formulated as:
\begin{equation}
    A(x) = f_a^a(h_\text{b}(x), h_\text{f}(x)),
\end{equation}
where the attack model $f_a^a$ leverages features from the biased model $h_\text{b}(x)$ and features from the fair model $h_\text{f}(x)$. By combining information from both models, the attack model $A(x)$ can make more accurate predictions about the sensitive subgroup information.

\subsection{Discussions}
The introduced fairness disparity based attack mechanism is designed to enhance the attack performance by leveraging predictions from both biased and fair models.
Unlike existing attack methods, this mitigates the risk of degraded performance in the trained attack model.
Our findings reveal that fairness interventions inadvertently introduce new privacy risks, making target models more vulnerable to membership inference attacks and attribute inference attacks.

\section{Experiments}
We now extensively evaluate our findings and the proposed method under diverse scenarios. We start by introducing the experiment settings.

\subsection{Settings}
\noindent \textbf{Datasets.} With the \textit{gender} attribute, we consider the following binary classifications: smiling predictions (T=s/S=g) with the CelebA dataset~\cite{CelebAMask-HQ}, race predictions (T=r/S=g) with the UTKFace dataset~\cite{geraldsutkface} and the FairFace dataset~\cite{karkkainen2021fairface}. As UTKFace and FairFace contain multiple racial subgroups, we first group them into \textit{White} and \textit{Others} and then obtain the binary subgroups.

\noindent \textbf{Training data.} For training data, we sample the data to skew the distribution with specified sensitive attributes to induce biased predictions. We set a highly imbalanced ratio of $9:1$ between the majority and minority groups (e.g., $90\%$ male data and $10\%$ female data). Meanwhile, we maintain a balanced distribution for the target learning (e.g., $50\%$ smiling and $50\%$ non-smiling). We also maintain a balanced distribution for the test data. In the ablation studies, we consider different imbalanced ratios to further verify the proposed method.

\noindent \textbf{Models.} For the target models, we utilize a 6-layer deep model comprising three consecutive CNN layers followed by three linear layers. We adopt the fair mixup operations from~\cite{mroueh2021fair,du2021fairness} to obtain fair models. We follow their implementations to apply the fairness intervention.

\subsection{Results with the \textit{gender} attribute}
Table~\ref{table:results_gender} presents the attack results with different attack methods and metrics. 
We integrate the proposed method with score-based MIAs ($\text{MIA}_\text{s}$), reference-based MIAs ($\text{MIA}_\text{l}$), the black-box AIAs ($\text{AIA}_\text{b}$), and the white-box AIAs ($\text{AIA}_\text{w}$). 
For $\text{MIA}_\text{l}$,
we report the TPR results at a low FPR value of $0.1\%$, following suggestions in~\cite{carlini2022membership}. 
The table shows that FD-MIA and FD-AIA outperform all existing attack methods with all cases and metrics. Notably, it achieves higher attack success on fair models than the biased ones. In contrast, the existing MIAs and AIAs perform worse on fair models. This reveals that the proposed FD-MIA and FD-AIA can effectively exploit model fairness disparities to improve attack performance, posing threats to model privacy.

\begin{figure}[t]
    \centering
    \subfloat[Score-based attacks \label{fig:lowFPR_score}]{
        \includegraphics[width=0.23\textwidth]{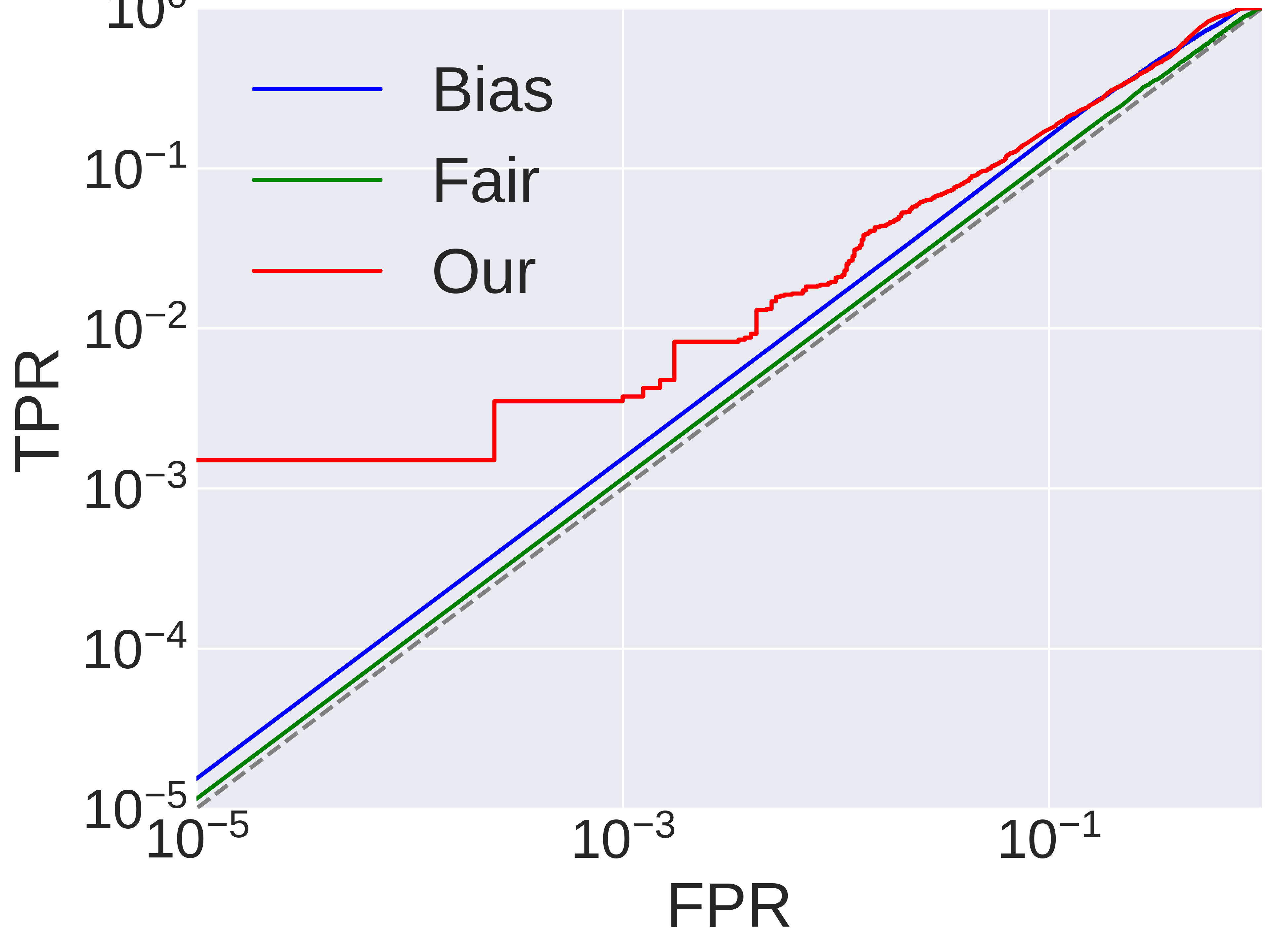}
    }
    \subfloat[LiRA attacks \label{fig:lowFPR_lira}]{
        \includegraphics[width=0.23\textwidth]{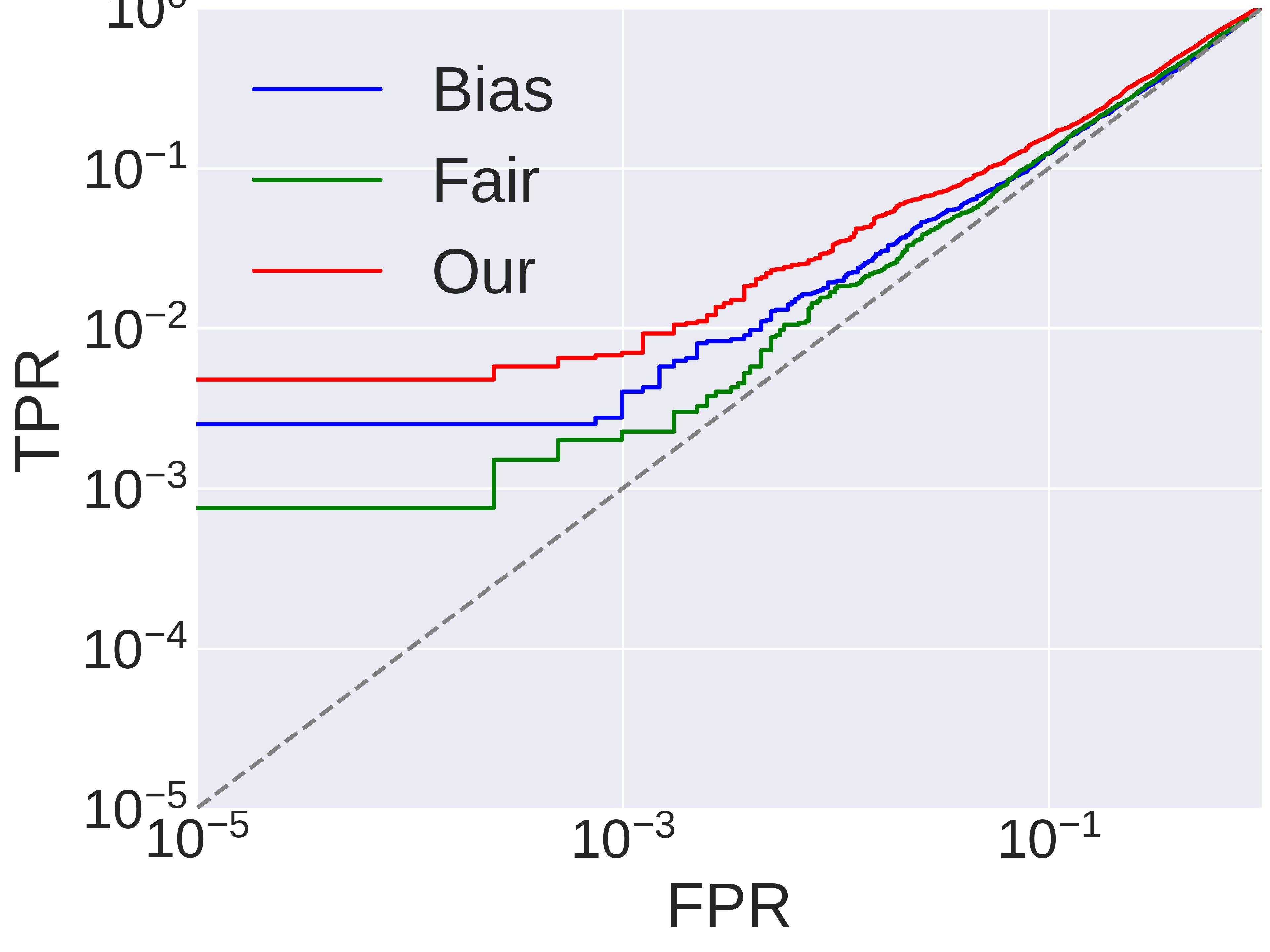}
    }
    \caption{Attack result comparisons in the low FPR region for (a) score-based attacks and (b) LiRA attacks.}
    \label{fig:dual_fpr}
\end{figure}

Specifically, in score-based attacks ($\text{MIA}_\text{s}$), FD-MIA outperformed others with the highest accuracy. Similar trends can be observed with the LiRA attacks ($\text{MIA}_\text{l}$). We further present the ROC curves for the CelebA case in Figure~\ref{fig:dual_fpr}. The figure further confirms the invalid attacks of the existing methods and the valid TPR results of FD-MIA. Moreover, from the table, we observe that the attacks achieved superior results on FairFace compared to other datasets. Meanwhile, FairFace exhibits a greater discrepancy in fairness between biased and fair models. We believe the enlarged discrepancy leads to enlarged prediction gaps, enabling more effective attacks. Additionally, we notice that score-based attacks perform better on accuracy and AUC, whereas LiRA achieves better TPR values. This aligns with the observations in~\cite{carlini2022membership} as LiRA is designed for efficient attacks at the low FPR. 

For AIAs, the table shows decreased attack performance with fair models when using existing methods. This is because fairness interventions aim to reduce prediction gaps across different demographic subgroups, making it more challenging to distinguish subgroups. Specifically, for CelebA, the accuracy of black-box AIAs drops from $83.4\%$ on biased models to $82.9\%$ on fair models, and white-box AIAs show a similar trend ($92.9\%$ to $91.6\%$).
In contrast, with FD-AIA, we see improved attack performance on fair models. For instance, on CelebA, FD-AIA achieves $87.6\%$ accuracy on fair models compared to $83.4\%$ on biased models for black-box attacks and $93.8\%$ vs $92.9\%$ for white-box attacks. This trend is consistent across all datasets, demonstrating that FD-AIA can effectively leverage the fairness-induced changes in the model to enhance attack success.

\begin{table*}[t]
\centering 
\caption{Target learning and attack results for the \textit{gender} attribute in (\%).}
  \begin{tabular}{c c c c c c c c c c}
  \toprule
  \multirow{2}{*}[-0.5ex]{Dataset}     &  \multirow{2}{*}[-0.5ex]{Models}  & \multicolumn{3}{c}{Target prediction results}                                & \multicolumn{4}{c}{Attack results}                      \\
                 \cmidrule(r){3-5} \cmidrule(l){6-9}
                      &  & $\text{Acc}_\text{t}$ $\uparrow$ & $\text{BA}$ $\downarrow$ & $\text{DEO}$ $\downarrow$ & $\text{MIA}_\text{s}$ (Acc) & $\text{MIA}_\text{l}$ (TPR)    & $\text{AIA}_\text{b}$  (Acc)   &  $\text{AIA}_\text{w}$ (Acc)    \\

    \cmidrule(r){1-2} \cmidrule(r){3-5} \cmidrule(l){6-9} 
            \multirow{3}{*}{\parbox{1.5cm}{\centering CelebA \\(T=s/S=g)}}
            &{$\text{Bias}$} & 87.6 ($\pm$0.0) &7.7 ($\pm$0.1) & 21.7 ($\pm$0.0) &59.8 ($\pm$0.0) & 0.6 ($\pm$0.1) & 56.5 ($\pm$0.5) & 83.4 ($\pm$0.5)\\
            &{$\text{Fair}$} &90.5 ($\pm$0.0) &2.5 ($\pm$0.9) &5.6 ($\pm$1.0) &53.2 ($\pm$1.1) &0.2 ($\pm$0.0) & 47.5 ($\pm$0.4) & 82.9 ($\pm$0.8) \\
            &{$\text{Our}$} &- &- &- &\textbf{60.6 ($\pm$0.2)} &\textbf{1.2 ($\pm$0.3)} & \textbf{75.2 ($\pm$0.2)} & \textbf{87.6 ($\pm$0.6)} \\
    \cmidrule(r){1-2} \cmidrule(r){3-5} \cmidrule(l){6-9}
            \multirow{3}{*}{\parbox{1.5cm}{\centering UTKFace \\(T=r/S=g)}}
            &{$\text{Bias}$} & 87.4 ($\pm$0.4) &3.6 ($\pm$0.3) &14.2 ($\pm$0.2)&58.5 ($\pm$0.1) &0.9 ($\pm$0.4) & 
            66.0 ($\pm$0.7) & 84.5 ($\pm$0.2)\\
            &{$\text{Fair}$} &89 ($\pm$0.5) &0.8 ($\pm$0.1) &6.3($\pm$0.2)&52.6 ($\pm$0.1) &0.7 ($\pm$0.3) & 
            57.0 ($\pm$0.5) & 84.1 ($\pm$0.1)\\
            &{$\text{Our}$} &-&-&- &\textbf{60.2 ($\pm$2.5)} &\textbf{1.7 ($\pm$0.3)} & \textbf{77.3 ($\pm$0.9)} & \textbf{85.9 ($\pm$0.1)}\\
    \cmidrule(r){1-2} \cmidrule(r){3-5} \cmidrule(l){6-9}
            \multirow{3}{*}{\parbox{1.5cm}{\centering FaceFace \\(T=r/S=g)}}
            &{$\text{Bias}$} & 87.2 ($\pm$0.0) &7.7 ($\pm$0.0) &22.2 ($\pm$0.4)&63.6 ($\pm$0.2) &1.3 ($\pm$0.5)  & 
            51.9 ($\pm$0.5) & 68.3 ($\pm$0.2)\\ 
            &{$\text{Fair}$} &87.6 ($\pm$0.1) &1.9 ($\pm$0.6) &3.9($\pm$0.2)&63.3 ($\pm$0.3) &0.9 ($\pm$0.1) &
            49.6 ($\pm$0.3) & 67.8 ($\pm$0.4)\\
            &{$\text{Our}$} &-&-&-&\textbf{65.2 ($\pm$0.1)} &\textbf{2.3 ($\pm$0.3)} & \textbf{53.4 ($\pm$0.9)} & \textbf{73.6 ($\pm$0.7)}\\ 
    \bottomrule
  \end{tabular}
  \label{table:results_gender}
\end{table*}

\subsection{Results with other attributes} We further explore attacks with different attributes, including \textit{wavy hair} (T=s/S=h) and \textit{heavy makeup} (T=s/S=m) for CelebA, as well as \textit{race} (T=g/S=r) for UTKFace and FairFace. Table~\ref{table:other_attrs} presents the results. Once again, the proposed FD-MIA and FD-AIA outperform existing attack methods on all datasets and metrics, posing real privacy threats. Notably, it consistently achieves superior performance with varying accuracy, ranging from $51\%$ to $77\%$. The results illustrate the robustness of the proposed methods, highlighting their efficacy in real-world scenarios. Additionally, similar to previous results, the proposed attack methods achieve better attack performance on FairFace, likely due to the enlarged fairness discrepancy between fair and biased models. 

\begin{table*}[t]
\centering 
\caption{Attacks with different sensitive attributes and learning targets in (\%).}
  \begin{tabular}{c c c c c c c c c c}
  \toprule
  \multirow{2}{*}[-0.5ex]{Dataset}     &  \multirow{2}{*}[-0.5ex]{Models}  & \multicolumn{3}{c}{Target prediction results}                                & \multicolumn{4}{c}{Attack results}                      \\
                 \cmidrule(r){3-5} \cmidrule(l){6-9}
                      &  & $\text{Acc}_\text{t}$ $\uparrow$ & $\text{BA}$ $\downarrow$ & $\text{DEO}$ $\downarrow$ & $\text{MIA}_\text{s}$ (Acc) & $\text{MIA}_\text{l}$ (TPR)    & $\text{AIA}_\text{b}$  (Acc)   &  $\text{AIA}_\text{w}$ (Acc)    \\
    \cmidrule(r){1-2} \cmidrule(r){3-5} \cmidrule(l){6-9} 
            \multirow{3}{*}{\parbox{1.5cm}{\centering CelebA \\(T=s/S=h)}}
            &{$\text{Bias}$} & 89.9 ($\pm$0.1) &2.5 ($\pm$0.0) &10.4 ($\pm$0.1) &55.1 ($\pm$0.1) & 0.3 ($\pm$0.1) & 54.2 ($\pm$0.1) & 60.9 ($\pm$0.3)  \\
            &{$\text{Fair}$} &90.1 ($\pm$0.4) &0.9 ($\pm$0.2) &3.7($\pm$0.5) &52.6 ($\pm$0.1) &0.1 ($\pm$0.1) & 51.9 ($\pm$0.5) & 59.7 ($\pm$0.4) \\
            &{$\text{Our}$} &- &- &- &\textbf{55.4 ($\pm$0.5)} &\textbf{0.8 ($\pm$0.1)} &\textbf{57.1 ($\pm$0.6)} &\textbf{62.5 ($\pm$0.6)} \\
    \cmidrule(r){1-2} \cmidrule(r){3-5} \cmidrule(l){6-9} 
            \multirow{3}{*}{\parbox{1.5cm}{\centering CelebA \\(T=s/S=m)}}
            &{$\text{Bias}$} & 88.6 ($\pm$0.1) &3.5 ($\pm$0.0) &14.6 ($\pm$0.1) &57.4 ($\pm$0.1) & 0.4 ($\pm$0.1) & 57.1 ($\pm$0.1) & 75.6 ($\pm$0.3) \\
            &{$\text{Fair}$} &90.5 ($\pm$0.3) &0.8 ($\pm$0.1) &2.4($\pm$0.6) &53.1 ($\pm$0.3) &0.1 ($\pm$0.1) & 
            51.1 ($\pm$0.3) & 75.1 ($\pm$0.1) \\
            &{$\text{Our}$} &- &- &- &\textbf{59.6 ($\pm$0.2)} &\textbf{0.6 ($\pm$0.1)} &\textbf{69.1 ($\pm$0.1)} &\textbf{77.6 ($\pm$0.1)} \\
    \cmidrule(r){1-2} \cmidrule(r){3-5} \cmidrule(l){6-9}
            \multirow{3}{*}{\parbox{1.5cm}{\centering UTKFace \\(T=g/S=r)}}
            &{$\text{Bias}$} & 80.8 ($\pm$0.1) &8.8 ($\pm$0.6) &31.9 ($\pm$1.4)&64.0 ($\pm$1.3) &1.4 ($\pm$0.1) & 
            56.9 ($\pm$0.1) & 70.4 ($\pm$0.5)\\
            &{$\text{Fair}$} &86.3 ($\pm$0.4) &2.8 ($\pm$0.4) &14.3($\pm$0.4)&55.3 ($\pm$0.8) &0.9 ($\pm$0.1) &
            52.0 ($\pm$0.4) & 69.4 ($\pm$0.1) \\
            &{$\text{Our}$} &-&-&- &\textbf{66.7 ($\pm$0.1)} &\textbf{2.1 ($\pm$0.3)} & \textbf{64.3 ($\pm$0.1)} & \textbf{73.8 ($\pm$0.0)}\\
    \cmidrule(r){1-2} \cmidrule(r){3-5} \cmidrule(l){6-9}
            \multirow{3}{*}{\parbox{1.5cm}{\centering FaceFace \\(T=g/S=r)}}
            &{$\text{Bias}$} & 90.5 ($\pm$0.3) &12.5 ($\pm$0.7) &5.3 ($\pm$1.1) &75.5 ($\pm$1.7) &1.5 ($\pm$0.1)  & 61.0 ($\pm$0.2) & 64.9 ($\pm$0.4)\\ 
            &{$\text{Fair}$} &92.0 ($\pm$0.3) &5.1 ($\pm$2.0) &4.5($\pm$1.1) &73.2 ($\pm$0.9) &0.6 ($\pm$0.4) &
            51.7 ($\pm$0.3) &63.6 ($\pm$0.9)\\
            &{$\text{Our}$} &-&-&-&\textbf{77.0 ($\pm$0.3)} &\textbf{2.9 ($\pm$0.7)} & \textbf{65.7 ($\pm$0.6)} & \textbf{77.4 ($\pm$0.3)}\\ 
    \bottomrule
  \end{tabular}
  \label{table:other_attrs}
\end{table*}

\subsection{Ablation studies}
\begin{figure}[t]
\begin{center}
\includegraphics[width=.9\linewidth]{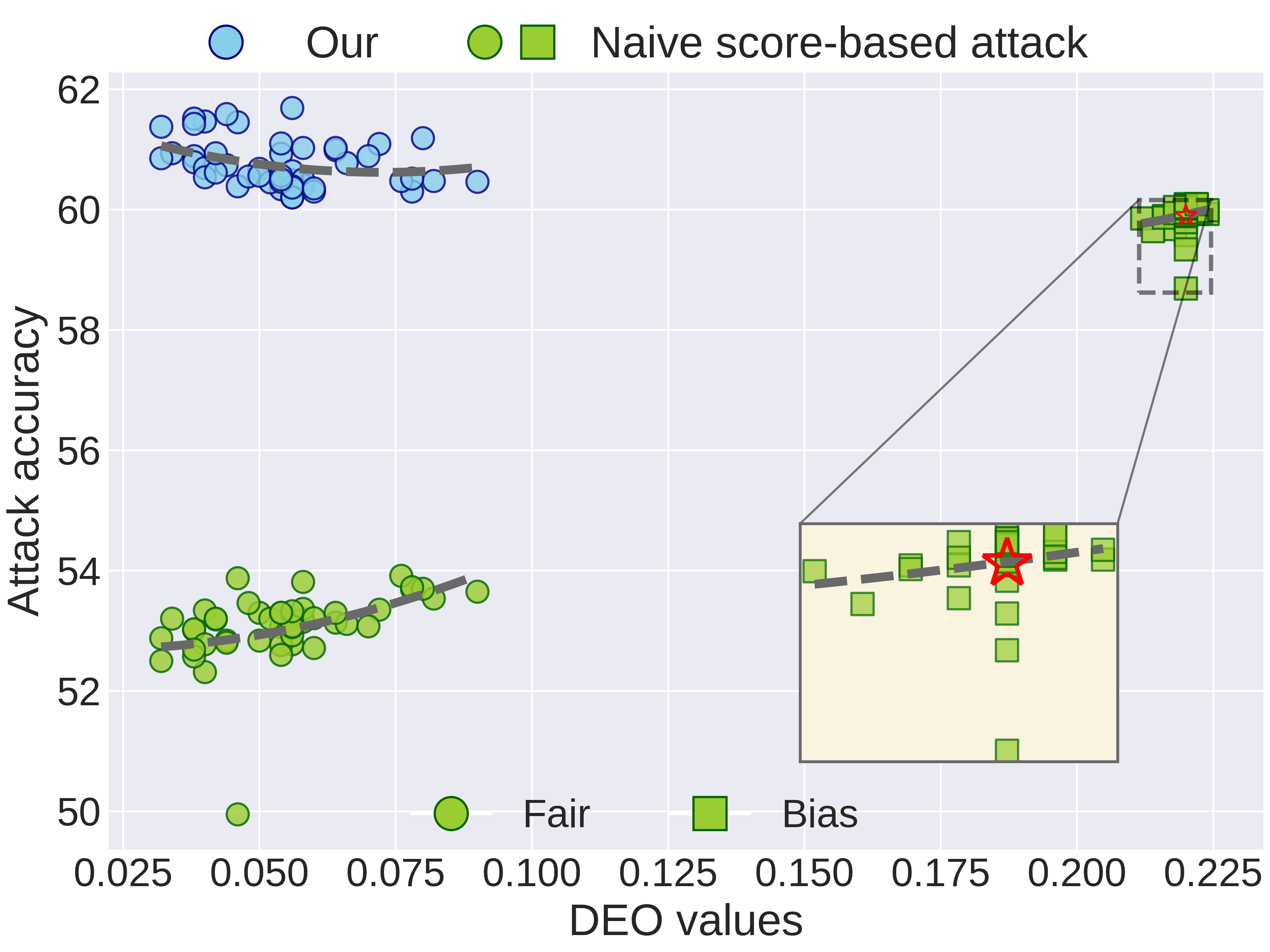}
\end{center}
\caption{Score-based MIAs with models of varying fairness levels. The \textit{red star} indicates the biased model.}
\label{fig:compare}
\end{figure}

\noindent \textbf{Results with varying fairness levels.} We attack models of different fairness performances. Here, we conduct score-based MIAs to evaluate the results. Specifically, we consider the case of CelebA (T=s/S=g) and conduct attacks on biased and fair models of different DEO values. Figure~\ref{fig:compare} presents the results. For FD-MIA, we utilize prediction results from multiple fair models and one biased one, which is indicated by a red star in the figure. We further adopt dashed gray lines to outline the trend. 

The figure illustrates that attack accuracy decreases for both biased and fair models as the DEO value decreases. The results indicate that models with stronger fairness interventions exhibit more robustness against existing MIAs. While achieving improved fairness, these models lower their confidence scores, making the attacks more challenging. In contrast, FD-MIA, which exploits discrepancies in fairness, achieves superior attack performance. Particularly, larger fairness discrepancies contribute to more powerful attacks. 

\begin{figure}[t]
\begin{center}
\includegraphics[width=.9\linewidth]{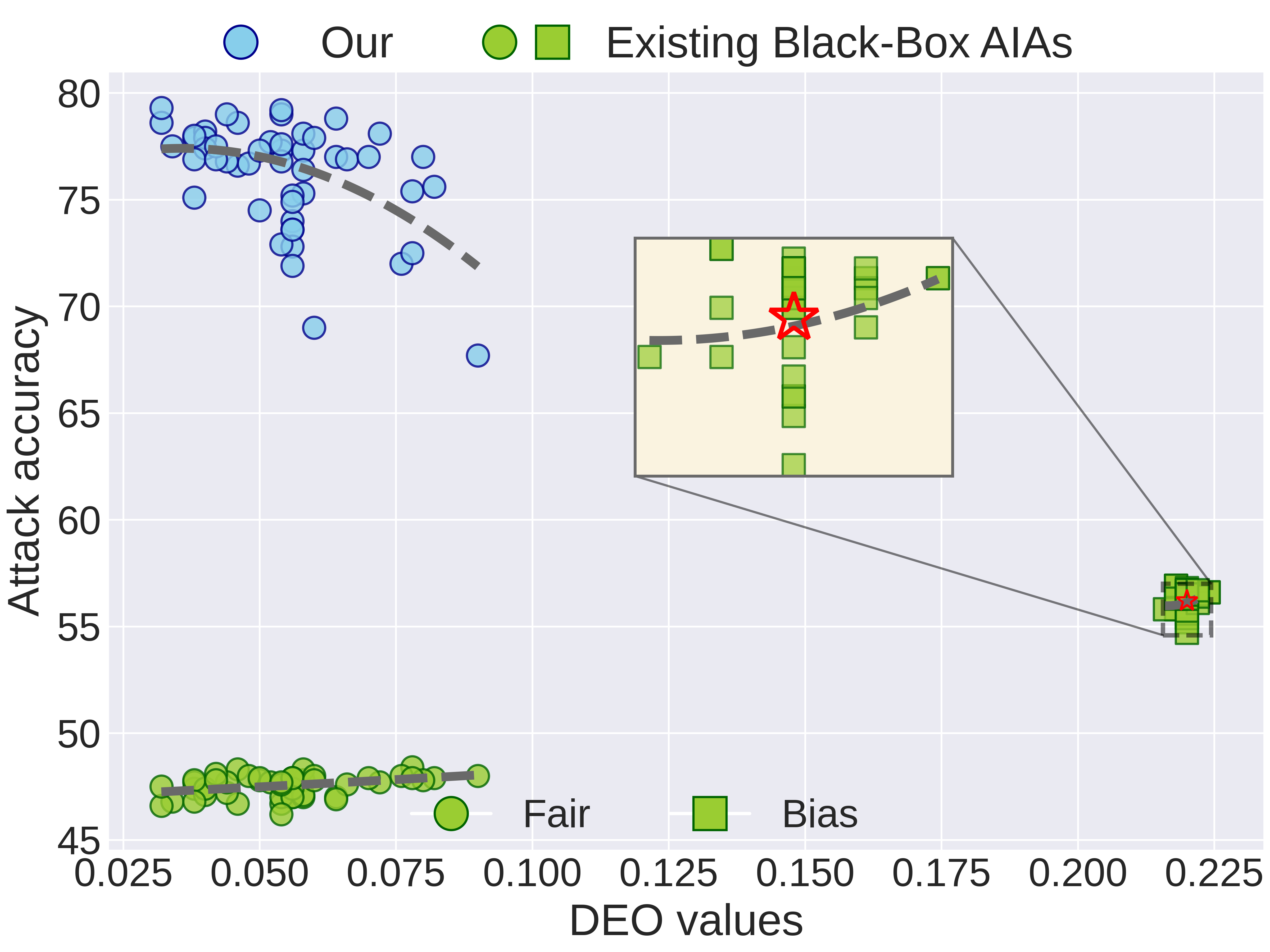}
\end{center}
\caption{Black-Box AIAs with models of varying fairness levels. The \textit{red star} indicates the biased model.}
\label{fig:aiacompare1}
\end{figure}

We further examine AIA results with different fair models in Figure~\ref{fig:aiacompare1}. Similarly, we observe that as the DEO value decreases, indicating improved fairness, the accuracy of traditional AIAs tends to decrease. This trend aligns with our previous observations that fairness interventions make it more difficult for standard attacks to infer sensitive attributes.
However, the FD-AIA method shows a different trend. As the fairness discrepancy between the biased and fair models increases (\textit{i.e.}, as the fair model's DEO decreases further from the biased model's DEO), the accuracy of FD-AIA improves. This is because FD-AIA leverages these fairness discrepancies to enhance its attack effectiveness.

\begin{figure}[t]
\begin{center}
\includegraphics[width=\linewidth]{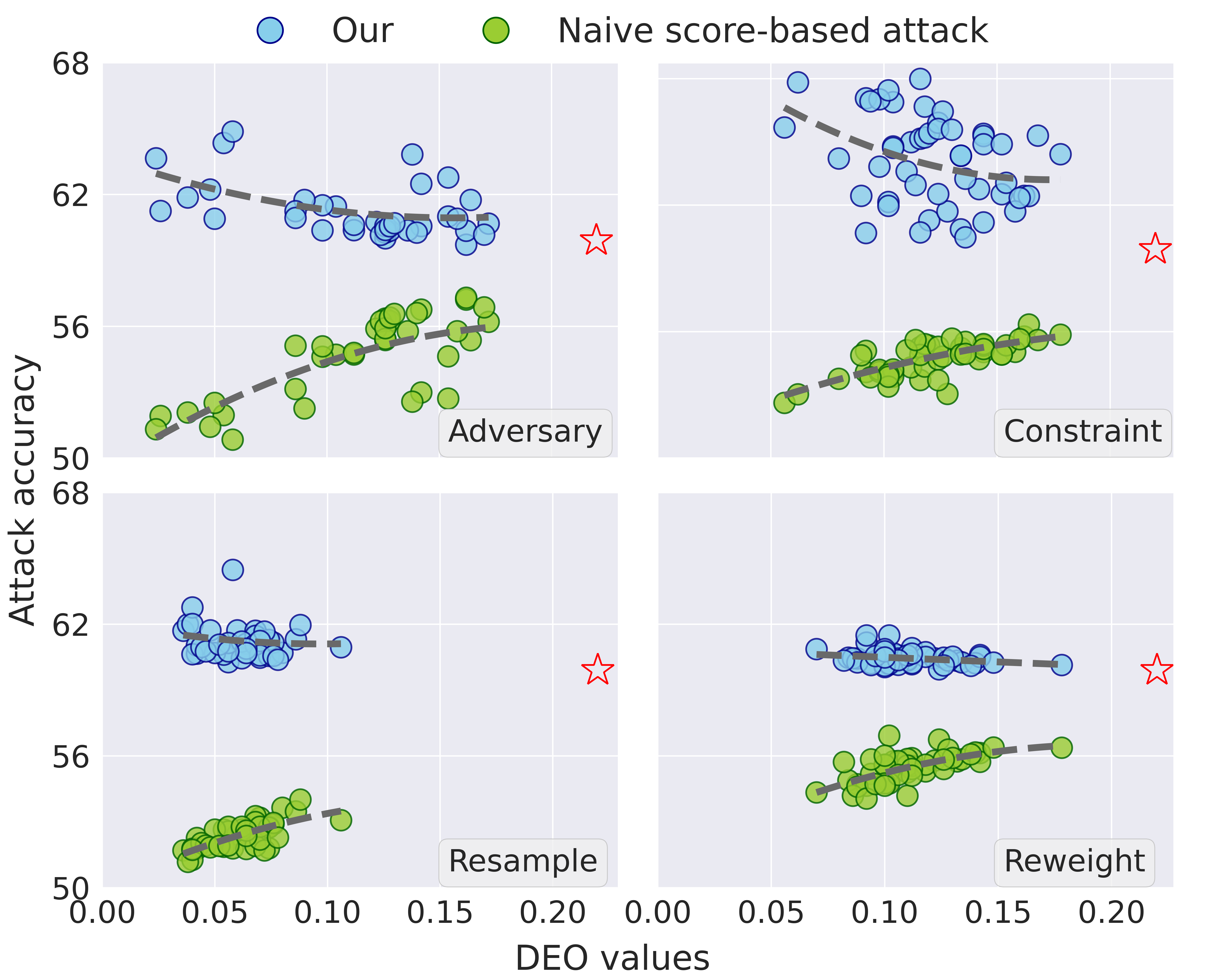}
\end{center}
\caption{Score-based MIAs on models with different fairness intervention methods.}
\label{fig:fair_methods}
\end{figure}

\noindent \textbf{Results with different fairness approaches.} We evaluate our findings with various fairness approaches, including data sampling, reweighting, adversarial training, and constraint-based approaches. In the experiments, we adopt the implementations of these approaches from~\cite{wangMitigatingBiasFace,han2023ffb}. 
Similarly, we focus on the case of CelebA (T=s/S=g), and Figure~\ref{fig:fair_methods} presents the results. The figure shows reduced DEO values after fairness interventions, indicating the effectiveness of these approaches.

For attack results, the naive score-based attacks exhibit degraded performance with fair models for all fairness approaches. The attack accuracy drops as the DEO values reduce. The results align with our previous findings, where fairness interventions introduce some robustness to MIAs. Notably, the drops are more pronounced with the adversarial training and constraint approaches. We believe this is due to the more substantial trade-offs between fairness and utility inherent to the approaches. 

\begin{figure}[t]
\begin{center}
\includegraphics[width=\linewidth]{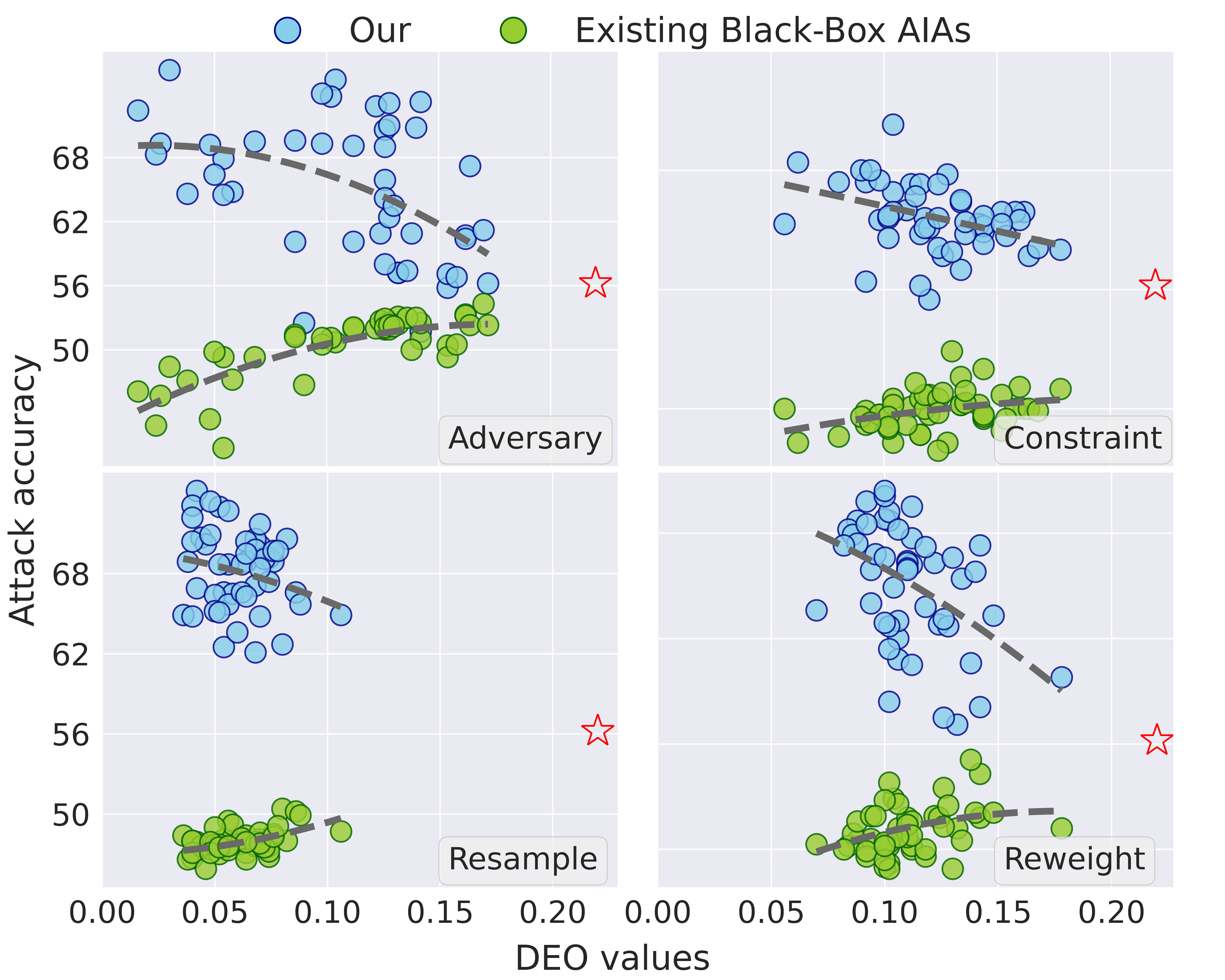}
\end{center}
\caption{Black-box AIAs on models with different fairness intervention methods.}
\label{fig:aiacompare2}
\end{figure}

In contrast, for all approaches, FD-MIA achieved higher attack accuracy with fair models compared to biased ones. Similarly, the attack performance improves when the fairness discrepancy enlarges as FD-MIA explores the prediction gaps. A similar trend can also be observed with AIAs in Figure~\ref{fig:aiacompare2}. These experiments demonstrate our findings and the proposed method with various representative fairness approaches. The results indicate fairness interventions can impose real privacy threats. 

\noindent \textbf{Results with different model structures.}
We assess the performance of the proposed method considering different model structures: 
ResNet18~\cite{He2016DeepRL} and VGG~\cite{Simonyan2015VeryDC}. Table~\ref{table:models} shows the results with score-based attacks. 
As consistently observed in our evaluations, the proposed
FD-MIA and FD-AIA outperform existing attack methods with all model structures. 
Notably, 
it achieves better attack performance with lighter model structures, 
such as ResNet18, 
compared to VGG. 
This can be attributed to the fact that lighter models are more susceptible to the influence of imbalanced data distributions, 
leading to more biased predictions. 
Consequently, 
this imbalance results in larger prediction gaps between member and non-member data, 
thereby contributing to enhanced attack performance. 

\begin{table}[t]
\centering 
\caption{Attack results with different model structures in (\%).}
  \begin{tabular}{c c c c}
  \toprule

   \multirow{2}{*}{Structures}     &  \multirow{2}{*}{Models}   & \multicolumn{2}{c}{Attack accuracy}                      \\
                 \cmidrule{3-4}
                 & & $\text{MIA}_\text{s}$ & $\text{AIA}_\text{b}$    \\ 
    \midrule 
            \multirow{3}{*}{Light CNN ~\cite{LWHSZBCFZ22}}
            &{$\text{Bias}$}&59.8 ($\pm$0.0) &56.5 ($\pm$0.5)   \\
            &{$\text{Fair}$}&53.2 ($\pm$1.1) &47.5 ($\pm$0.4)   \\
            &{$\text{Our}$}&\textbf{60.6 ($\pm$0.2)} &\textbf{75.2 ($\pm$0.2)} \\
    \midrule
            \multirow{3}{*}{ResNet18 ~\cite{He2016DeepRL}}
            &{$\text{Bias}$}&59.6 ($\pm$0.6) &55.8 ($\pm$0.2)  \\
            &{$\text{Fair}$}&54.2 ($\pm$0.1) &48.4 ($\pm$0.6)  \\
            &{$\text{Our}$}&\textbf{64.5 ($\pm$0.0)} &\textbf{74.9 ($\pm$0.4)}\\
    \midrule
            \multirow{3}{*}{VGG ~\cite{Simonyan2015VeryDC}}
            &{$\text{Bias}$} &55.2 ($\pm$0.2) &54.1 ($\pm$0.2)  \\
            &{$\text{Fair}$} &52.2 ($\pm$0.8) &45.2 ($\pm$0.3)  \\
            &{$\text{Our}$}  &\textbf{59.6 ($\pm$0.2)} & \textbf{72.2 ($\pm$0.3)} \\
    \bottomrule
  \end{tabular}
  \label{table:models}
\end{table}

\noindent \textbf{Attacks with varying skewed distributions.}
We further consider varying skewed data distributions for the considered sensitive attribute. Specifically, we consider smiling classifications with the CelebA dataset considering \textit{gender} as the sensitive attribute. We then sample the data randomly and set imbalanced ratios between the majority and minority subgroups with different values ranging from $0.95$ to $0.75$. Table~\ref{table:table_dist} presents the results for target learning and the attack results. All results indicate that fairness interventions tend to introduce some robustness to MIAs and AIAs. However, with the proposed attack method, fair models can compromise model privacy with superior attack results. The results are consistent with our findings.

\begin{table*}[t]
\centering 
\caption{Attack results for different skewed distributions.}
  \begin{tabular}{c c c c c c c c c c}
  \toprule
  \multirow{2}{*}[-0.5ex]{Distributions}     &  \multirow{2}{*}[-0.5ex]{Models}  & \multicolumn{3}{c}{Target prediction results}                                & \multicolumn{4}{c}{Attack results}                      \\
                 \cmidrule(r){3-5} \cmidrule(l){6-9}
                      &  & $\text{Acc}_\text{t}$ $\uparrow$ & $\text{BA}$ $\downarrow$ & $\text{DEO}$ $\downarrow$ & $\text{MIA}_\text{s}$ (Acc) & $\text{MIA}_\text{l}$ (TPR)    & $\text{AIA}_\text{b}$  (Acc)   &  $\text{AIA}_\text{w}$ (Acc)    \\
    \cmidrule(r){1-2} \cmidrule(r){3-5} \cmidrule(l){6-9} 
            \multirow{3}{*}{0.95}
            &{$\text{Bias}$} & 82.7 ($\pm$0.0) &9.6 ($\pm$0.0) &31.0 ($\pm$0.0) &63.8 ($\pm$0.4) & 0.0 ($\pm$0.0) & 60.1 ($\pm$0.3) & 86.3 ($\pm$0.1)  \\
            &{$\text{Fair}$} &89.3 ($\pm$0.3) &3.3 ($\pm$0.8) &9.2($\pm$1.1) &54.9 ($\pm$0.4) &0.2 ($\pm$0.0) & 50.9 ($\pm$0.6) & 85.1 ($\pm$0.5) \\
            &{$\text{Our}$} &- &- &- &\textbf{64.5 ($\pm$0.3)} &\textbf{0.2 ($\pm$0.0)} &\textbf{79.4 ($\pm$0.6)} &\textbf{89.1 ($\pm$0.6)} \\
    \cmidrule(r){1-2} \cmidrule(r){3-5} \cmidrule(l){6-9} 
            \multirow{3}{*}{0.9}
            &{$\text{Bias}$} & 87.6 ($\pm$0.0) &7.7 ($\pm$0.1) &21.7 ($\pm$0.0) &59.8 ($\pm$0.0) & 0.0 ($\pm$0.0) & 58.2 ($\pm$0.1) & 86.1 ($\pm$0.2) \\
            &{$\text{Fair}$} &90.5 ($\pm$0.0) &2.5 ($\pm$0.9) &5.6($\pm$1.0) &53.2 ($\pm$1.1) &0.0 ($\pm$0.0) & 
            50.2 ($\pm$0.4) & 85.2 ($\pm$0.5) \\
            &{$\text{Our}$} &- &- &- &\textbf{60.6 ($\pm$0.2)} &\textbf{0.3 ($\pm$0.1)} &\textbf{78.9 ($\pm$0.6)} &\textbf{88.6 ($\pm$0.8)} \\
    \cmidrule(r){1-2} \cmidrule(r){3-5} \cmidrule(l){6-9}
            \multirow{3}{*}{0.85}
            &{$\text{Bias}$} & 88.6 ($\pm$0.3) &4.8 ($\pm$0.2) &17.0 ($\pm$0.7)&57.6 ($\pm$0.7) &0.0 ($\pm$0.0) & 
            56.9 ($\pm$0.1) & 84.2 ($\pm$0.5)\\
            &{$\text{Fair}$} &90.1 ($\pm$0.5) &1.7 ($\pm$0.8) &3.8($\pm$0.8)&54.1 ($\pm$0.3) &0.0 ($\pm$0.0) &
            83.6 ($\pm$0.9) & 84.1 ($\pm$0.6) \\
            &{$\text{Our}$} &-&-&- &\textbf{60.3 ($\pm$0.1)} &\textbf{0.3 ($\pm$0.1)} & \textbf{76.3 ($\pm$0.7)} & \textbf{88.1 ($\pm$0.5)}\\
    \cmidrule(r){1-2} \cmidrule(r){3-5} \cmidrule(l){6-9}
            \multirow{3}{*}{0.8}
            &{$\text{Bias}$} & 88.1 ($\pm$0.3) &5.0 ($\pm$0.3) &12.4 ($\pm$0.6) &58.7 ($\pm$0.8) &0.0 ($\pm$0.0)  & 56.4 ($\pm$0.3) & 83.4 ($\pm$0.4)\\ 
            &{$\text{Fair}$} &90.5 ($\pm$0.3) &1.9 ($\pm$0.6) &4.1($\pm$1.1) &54.2 ($\pm$0.2) &0.0 ($\pm$0.0) &
            47.6 ($\pm$0.3) &82.6 ($\pm$0.3)\\
            &{$\text{Our}$} &-&-&-&\textbf{59.5 ($\pm$0.2)} &\textbf{0.2 ($\pm$0.0)} & \textbf{75.3 ($\pm$0.4)} & \textbf{87.4 ($\pm$0.6)}\\ 
    \cmidrule(r){1-2} \cmidrule(r){3-5} \cmidrule(l){6-9}
            \multirow{3}{*}{0.75}
            &{$\text{Bias}$} & 89.1 ($\pm$0.3) &5.0 ($\pm$0.3) &12.4 ($\pm$0.6) &58.7 ($\pm$0.8) &0.0 ($\pm$0.0)  & 56.0 ($\pm$0.2) & 83.4 ($\pm$0.1)\\ 
            &{$\text{Fair}$} &90.1 ($\pm$0.1) &0.8 ($\pm$0.3) &2.8($\pm$1.2) &53.8 ($\pm$0.3) &0.0 ($\pm$0.0) &
            47.1 ($\pm$0.3) &82.0 ($\pm$0.4)\\
            &{$\text{Our}$} &-&-&-&\textbf{58.3 ($\pm$0.7)} &\textbf{0.3 ($\pm$0.1)} & \textbf{75.3 ($\pm$0.5)} & \textbf{87.4 ($\pm$0.9)}\\ 
    \cmidrule(r){1-2} \cmidrule(r){3-5} \cmidrule(l){6-9}
            \multirow{3}{*}{0.7}
            &{$\text{Bias}$} & 92.0 ($\pm$0.3) &1.9 ($\pm$0.0) &4.8 ($\pm$0.1) &56.9 ($\pm$1.4) &0.0 ($\pm$0.0)  & 55.1 ($\pm$0.1) & 83.6 ($\pm$0.3)\\ 
            &{$\text{Fair}$} &91.4 ($\pm$0.4) &0.9 ($\pm$0.0) &2.7($\pm$0.6) &53.9 ($\pm$0.3) &0.0 ($\pm$0.0) &
            47.0 ($\pm$0.2) &82.2 ($\pm$0.1)\\
            &{$\text{Our}$} &-&-&-&\textbf{57.3 ($\pm$0.3)} &\textbf{0.3 ($\pm$0.0)} & \textbf{75.3 ($\pm$0.3)} & \textbf{87.2 ($\pm$0.2)}\\ 
    \bottomrule
  \end{tabular}
  \label{table:table_dist}
\end{table*}

\subsection{Discussions}
Our experimental results, as presented in the previous tables, show that the proposed FD-MIA and FD-AIA methods achieved improvements in attack performance, albeit modest in some cases. This is because we deliberately chose fair models that maintained relatively high accuracy levels. This decision was motivated by practical considerations, as models with severe accuracy degradation are less likely to be deployed in real-world scenarios. However, this choice resulted in smaller prediction gaps between the fair and biased models, which in turn limited the potential for our attack methods to exploit these differences. Despite these constraints, our findings in Figures ~\ref{fig:compare} and ~\ref{fig:fair_methods} reveal an important trend. The attack performance of FD-MIA and FD-AIA can be significantly enhanced when using fairer models that exhibit more substantial drops in accuracy or when employing fair methods that demonstrate more significant fairness-accuracy trade-offs.

\section{Mitigations}
Previous evaluations have demonstrated that the proposed method compromises model privacy performance. 
This section outlines two potential defense mechanisms to mitigate privacy leakage from the proposed method.

\subsection{Restricting information access} 
This method involves limiting the adversary's access to the information required for the attack methods. Specifically, we consider the following restrictions:
\begin{itemize}
    \item \textbf{Label-only access}: Only providing predicted labels without confidence scores or other intermediate outputs.
    \item \textbf{Fair model isolation}: Publishing only the prediction results from fair models, preventing adversaries from obtaining the prediction discrepancies that FD-MIA and FD-AIA exploit.
    \item \textbf{Prediction truncation}: Limiting the precision of confidence scores by rounding or truncation.
\end{itemize}

While effective, these approaches come with trade-offs: First, restricting to label-only access may limit the utility of deployed models, such as risk assessment systems requiring probability scores, medical diagnostics where confidence levels guide treatment decisions, or recommendation systems that rank items based on prediction scores. Second, fair model isolation may not be feasible when both models are part of a model evolution timeline, as organizations typically maintain version histories for audit purposes and compliance requirements that mandate preserving model histories. Third, users requiring high-precision outputs for decision support or quality control systems may find truncation unacceptable for their applications.

\subsection{Differential privacy} 
Differential privacy (DP)~\cite{dwork2006calibrating} imposes a constraint on the ability to distinguish between two neighbouring datasets that differ by only a single data sample, and research has shown that DP can effectively mitigate MIAs and AIAs. DP-SGD~\cite{abadi2016deep} is effective against our proposed attacks as it targets the vulnerability exploited by our attacks: the model's ability to memorize training data. By adding noise to gradients during training, DP-SGD limits how precisely the model can fit individual training data. This reduces the prediction discrepancies between different subgroups that our attacks exploit.

\noindent \textbf{Experimental results.}
We utilize the differentially private stochastic gradient descent (DP-SGD)~\cite{abadi2016deep} for attacks considering the results of CelebA (T=s/S=g) in Table~\ref{table:score_results}.
Table~\ref{table:dp_results} shows the results, 
where we compare the attack performance with DP noise between the proposed methods and existing ones (the score-based attacks $s$ and the LiRA attacks $l$). 
The results show lower accuracy results than the original attacks, indicating the effectiveness of the defense methods.
Moreover, 
with the same amount of noise,
our attacks ($\text{Our}_s$, $\text{Our}_l$) achieve higher attack performance than the others, indicating that the proposed models require more DP noise to attain comparable levels of defense performance. 
The results show that the proposed methods are more effective in attacks than the existing approaches. 

\begin{figure}[t]
\begin{center}
\includegraphics[width=.8\linewidth]{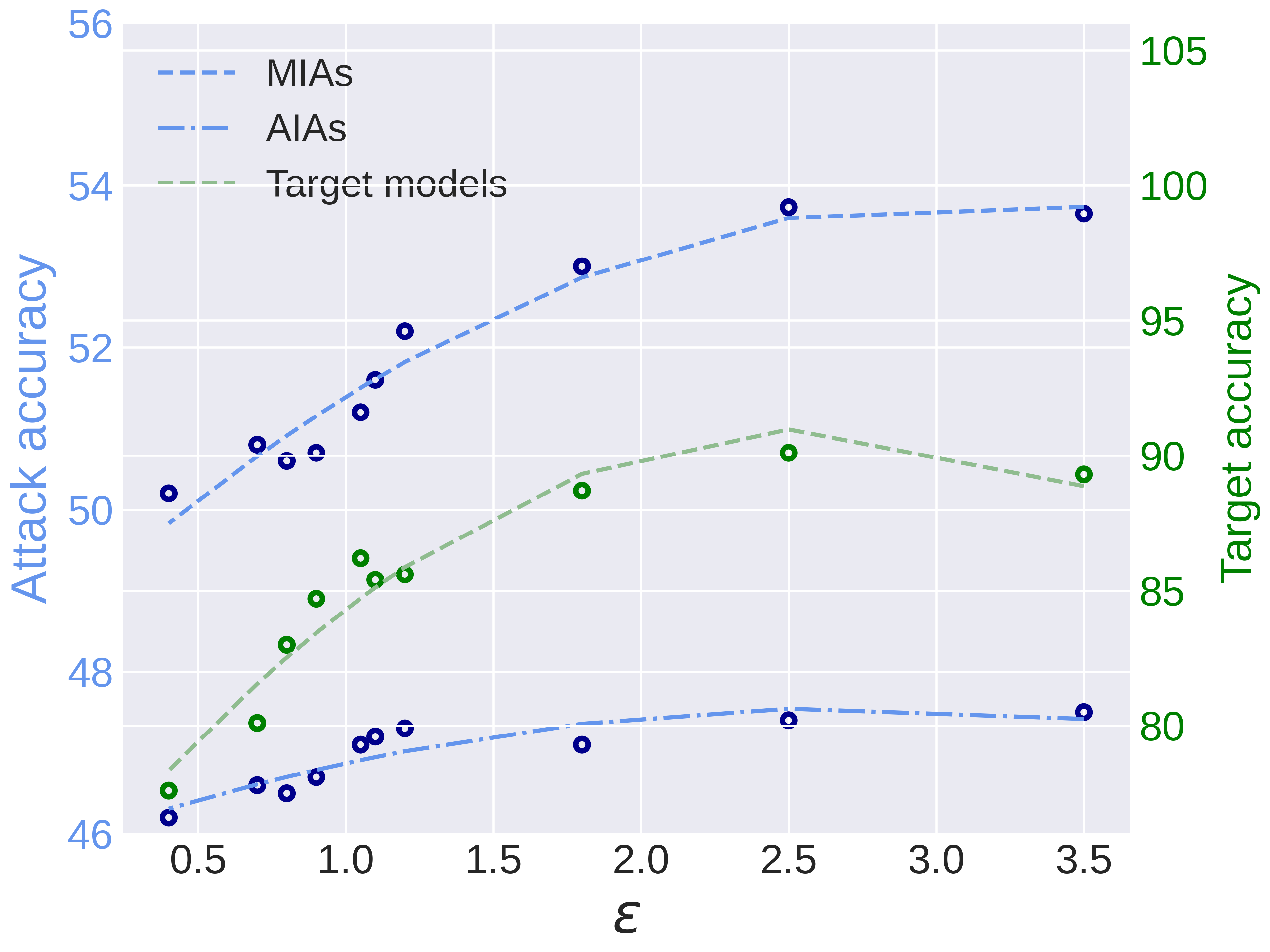}
\end{center}
\caption{DP-SGD results for different values of $\epsilon$. We compare accuracy results for target models and attack models.}
\label{fig:dp_results}
\end{figure}

Figure~\ref{fig:dp_results} illustrates the defense results with different values of the DP budget $\epsilon$.
In the figure, we report the accuracy results for the target predictions and the attack models. As shown in the figure, the DP noise will lead to \textbf{decreased} accuracy of learning targets. With smaller values of $\epsilon$, more noise will be injected during the training, leading to inferior attack performance but lower prediction performance. The figure demonstrates the trade-offs between privacy defense and model utility. The results also indicate that, with careful tuning of the noise budget $\epsilon$, DP-SGD can prevent privacy leaks from fairness-enforced models while maintaining performance on main predictions.

\begin{table}[t]
\centering 
\caption{DP-SGD results with $\delta=10^{-5}, \epsilon=0.85$ in (\%).}
  \begin{tabular}{c c c c c}
  \toprule
  \multirow{2}{*}[-0.5ex]{Models}  & \multicolumn{4}{c}{Attack results}                      \\
  \cmidrule(l){2-5}
                      & $\text{MIA}_\text{s}$ (Acc)& $\text{MIA}_\text{l}$ (TPR)   & $\text{AIA}_\text{b}$ (Acc)    &  $\text{AIA}_\text{w}$ (Acc)   \\
    \cmidrule(r){1-1} \cmidrule(r){2-5}
            
            {$\text{Fair}$} &50.8 ($\pm$0.3) &0.1 ($\pm$0.1) & 46.5 ($\pm$0.3) & 64.3 ($\pm$0.5) \\
            {$\text{Our}$} &53.4 ($\pm$0.4) &0.1 ($\pm$0.1) & 57.3 ($\pm$0.2) & 67.1 ($\pm$0.4) \\
    \bottomrule
  \end{tabular}
  \label{table:dp_results}
\end{table}

\subsection{Discussions}
Our analyses show that information access restriction methods provide a straightforward approach to mitigate FD-MIA and FD-AIA attacks. Particularly, fair model isolation effectively prevents adversaries from obtaining the prediction pairs essential for these attacks. While label-only access offers strong protection, it substantially reduces the model's usefulness for applications requiring confidence scores. Prediction truncation presents a middle ground, offering moderate protection with less utility impact.

Differential privacy methods provide the strongest theoretical guarantees against these attacks. With carefully chosen privacy budgets ($\epsilon \leq 1.0$), DP-SGD maintains reasonable model performance while significantly reducing attack success rates. This approach introduces a trade-off between privacy protection and model utility, making it suitable for high-sensitivity applications. 

The choice between these defense strategies depends on specific deployment requirements. Information restriction methods are easier to implement but more limiting for applications, while differential privacy offers stronger guarantees but requires more complex implementation and potentially greater utility sacrifices. For applications with strict privacy requirements, combining both approaches may provide the most comprehensive protection.

\section{Discussions}
\noindent \textbf{Computational cost.} While our proposed attack methods demonstrate improved effectiveness over existing approaches, they require more computational cost compared to conventional single-model attacks. Specifically, for the inference phase, the proposed attack methods require additional layers to process inputs from both biased and fair models. Similarly, during the training phase, the attacks need shadow models mimicking both the biased and fair target models. This doubles the training requirement compared to a single-model attack pipeline.

\noindent \textbf{Real-world feasibility.} The proposed FD-MIA and FD-AIA methods require predictions from both biased and fair models. The practical feasibility depends on the adversary's ability to access predictions from both biased and fair models. We identify several realistic scenarios. For example, as organizations continuously improve their models to address fairness concerns due to regulatory requirements, an adversary can record predictions from different versions of deployed models over time. Alternatively, the attacker could also deliberately report bias, compelling the model owner to implement fairness interventions. In collaborative ML environments where multiple stakeholders participate in model development, different versions of models (including biased and fair variants) might be accessible to participants, inadvertently providing information that could be exploited.

\section{Limitations}
While our study provides empirical evidence regarding privacy risks in fairness-enhanced machine learning models, there are limitations.

\noindent \textbf{Theoretical analyses.} Our focus lies in exploring and empirically measuring privacy leakage after enforcing fairness enhancement methods. We have not presented a comprehensive theoretical explanation of why fairness interventions can create exploitable prediction gaps. We have found that fairness methods adjust prediction scores differently for different subgroups. This results in prediction shifts that our FD-MIA and FD-AIA methods exploit. However, there is a broader theoretical question regarding whether such distribution shifts are tied to fairness enforcement or are contingent on specific optimization objectives. Future work might build on existing analytical frameworks to characterize these phenomena: explain how, when, and why fairness mechanisms may inadvertently reveal sensitive information.

\noindent \textbf{Fairness definition scope.}
Our study focuses on group fairness metrics (e.g., demographic parity or equalized odds). We find that fairness improvements do not necessarily increase privacy risks with naive attack methods. However, alternative fairness notions may lead to different privacy outcomes. Indeed, recent studies report trade-offs between fairness and privacy in various settings. For instance, Zhang et al.~\cite{zhang2024unraveling} explore how enforcing individual fairness in Graph Neural Networks can heighten privacy vulnerabilities. These findings underscore the need to evaluate fairness–privacy interactions across diverse fairness formulations.

\noindent \textbf{Scope and future directions.}
Our experiments focus on binary classification tasks. Although this setup offers a clear starting point for analyzing how fairness methods interact with privacy, real-world pipelines often involve more complex settings. For example, future research could extend to more complex model architectures, such as large language models, more applications in domains such as healthcare and finance, and alternative attack vectors such as model-stealing attacks.

\section{Conclusions}
This paper presents a comprehensive analysis of the interplay between algorithmic fairness methods and privacy vulnerabilities against membership and attribute inference attacks. Our extensive experiments across three datasets reveal that fairness interventions do not necessarily compromise model privacy when evaluated with existing MIA and AIA methods. However, we find that current attack methods are inadequate for fully assessing privacy leakage in fair models due to performance trade-offs and model degradation issues.
Motivated by these observations, we propose FD-MIA and FD-AIA novel attack methods tailored for fair models. These approaches exploit prediction disparities between original and fair models, consistently outperforming existing attacks and uncovering previously overlooked privacy risks in fair models.
Our findings underscore the need for a holistic approach to responsible AI system design that simultaneously addresses fairness and privacy concerns. The challenge of developing trustworthy systems that optimally balance these competing objectives remains an important area for future research.


{
\bibliographystyle{IEEEtran}
\bibliography{bib.bib}

\begin{thebibliography}{10}
\providecommand{\url}[1]{#1}
\csname url@samestyle\endcsname
\providecommand{\newblock}{\relax}
\providecommand{\bibinfo}[2]{#2}
\providecommand{\BIBentrySTDinterwordspacing}{\spaceskip=0pt\relax}
\providecommand{\BIBentryALTinterwordstretchfactor}{4}
\providecommand{\BIBentryALTinterwordspacing}{\spaceskip=\fontdimen2\font plus
\BIBentryALTinterwordstretchfactor\fontdimen3\font minus
  \fontdimen4\font\relax}
\providecommand{\BIBforeignlanguage}[2]{{%
\expandafter\ifx\csname l@#1\endcsname\relax
\typeout{** WARNING: IEEEtran.bst: No hyphenation pattern has been}%
\typeout{** loaded for the language `#1'. Using the pattern for}%
\typeout{** the default language instead.}%
\else
\language=\csname l@#1\endcsname
\fi
#2}}
\providecommand{\BIBdecl}{\relax}
\BIBdecl

\bibitem{brown2020language}
T.~Brown, B.~Mann, N.~Ryder, M.~Subbiah, J.~D. Kaplan, P.~Dhariwal,
  A.~Neelakantan, P.~Shyam, G.~Sastry, A.~Askell \emph{et~al.}, ``Language
  models are few-shot learners,'' \emph{Advances in neural information
  processing systems}, vol.~33, pp. 1877--1901, 2020.

\bibitem{kirillov2023segment}
A.~Kirillov, E.~Mintun, N.~Ravi, H.~Mao, C.~Rolland, L.~Gustafson, T.~Xiao,
  S.~Whitehead, A.~C. Berg, W.-Y. Lo \emph{et~al.}, ``Segment anything,''
  \emph{arXiv preprint arXiv:2304.02643}, 2023.

\bibitem{mehrabi2021survey}
N.~Mehrabi, F.~Morstatter, N.~Saxena, K.~Lerman, and A.~Galstyan, ``A survey on
  bias and fairness in machine learning,'' \emph{ACM Computing Surveys (CSUR)},
  pp. 1--35, 2021.

\bibitem{chang2021privacy}
H.~Chang and R.~Shokri, ``On the privacy risks of algorithmic fairness,'' in
  \emph{2021 IEEE European Symposium on Security and Privacy (EuroS\&P)}, 2021,
  pp. 292--303.

\bibitem{shokri2017membership}
R.~Shokri, M.~Stronati, C.~Song, and V.~Shmatikov, ``Membership inference
  attacks against machine learning models,'' in \emph{2017 IEEE symposium on
  security and privacy (SP)}, 2017, pp. 3--18.

\bibitem{LWHSZBCFZ22}
Y.~Liu, R.~Wen, X.~He, A.~Salem, Z.~Zhang, M.~Backes, E.~D. Cristofaro,
  M.~Fritz, and Y.~Zhang, ``{ML-Doctor: Holistic Risk Assessment of Inference
  Attacks Against Machine Learning Models},'' in \emph{{USENIX Security
  Symposium (USENIX Security)}}, 2022, pp. 4525--4542.

\bibitem{carlini2022membership}
N.~Carlini, S.~Chien, M.~Nasr, S.~Song, A.~Terzis, and F.~Tramer, ``Membership
  inference attacks from first principles,'' in \emph{2022 IEEE Symposium on
  Security and Privacy (SP)}, 2022, pp. 1897--1914.

\bibitem{Huantian2024}
H.~Tian, G.~Zhang, B.~Liu, T.~Zhu, M.~Ding, and W.~Zhou, ``When fairness meets
  privacy: Exploring privacy threats in fair binary classifiers via membership
  inference attacks,'' in \emph{International Joint Conference on Artificial
  Intelligence (IJCAI)}, 2024.

\bibitem{Zemel:2013wz}
R.~S. Zemel, L.~Y. Wu, K.~Swersky, T.~Pitassi, and C.~Dwork, ``Learning fair
  representations,'' in \emph{International Conference on Machine Learning},
  2013.

\bibitem{Manisha2020FNNCAF}
P.~Manisha and S.~Gujar, ``Fnnc: Achieving fairness through neural networks,''
  in \emph{International Joint Conference on Artificial Intelligence (IJCAI)},
  2020.

\bibitem{Xu_2021_CVPR}
X.~Xu, Y.~Huang, P.~Shen, S.~Li, J.~Li, F.~Huang, Y.~Li, and Z.~Cui,
  ``Consistent instance false positive improves fairness in face recognition,''
  in \emph{Proceedings of the IEEE/CVF Conference on Computer Vision and
  Pattern Recognition (CVPR)}, June 2021, pp. 578--586.

\bibitem{NEURIPS2021_fc2e6a44}
H.~C. Bendekgey and E.~Sudderth, ``Scalable and stable surrogates for flexible
  classifiers with fairness constraints,'' in \emph{Advances in Neural
  Information Processing Systems}, M.~Ranzato, A.~Beygelzimer, Y.~Dauphin,
  P.~Liang, and J.~W. Vaughan, Eds., vol.~34, 2021.

\bibitem{Tang_2023_CVPR}
P.~Tang, W.~Yao, Z.~Li, and Y.~Liu, ``Fair scratch tickets: Finding fair sparse
  networks without weight training,'' in \emph{Proceedings of the IEEE/CVF
  Conference on Computer Vision and Pattern Recognition (CVPR)}, 2023, pp.
  24\,406--24\,416.

\bibitem{truong2023fredom}
T.-D. Truong, N.~Le, B.~Raj, J.~Cothren, and K.~Luu, ``Fredom: Fairness domain
  adaptation approach to semantic scene understanding,'' in \emph{Proceedings
  of the IEEE/CVF Conference on Computer Vision and Pattern Recognition
  (CVPR)}, 2023.

\bibitem{cruz2023fairgbm}
A.~Cruz, C.~G. Bel{\'e}m, J.~Bravo, P.~Saleiro, and P.~Bizarro, ``Fair{GBM}:
  Gradient boosting with fairness constraints,'' in \emph{The Eleventh
  International Conference on Learning Representations}, 2023.

\bibitem{jung2023reweighting}
S.~Jung, T.~Park, S.~Chun, and T.~Moon, ``Re-weighting based group fairness
  regularization via classwise robust optimization,'' in \emph{The Eleventh
  International Conference on Learning Representations}, 2023.

\bibitem{xu2020BeRobustBe}
H.~Xu, X.~Liu, Y.~Li, A.~Jain, and J.~Tang, ``To be robust or to be fair:
  Towards fairness in adversarial training,'' in \emph{International Conference
  on Machine Learning}.\hskip 1em plus 0.5em minus 0.4em\relax PMLR, 2021, pp.
  11\,492--11\,501.

\bibitem{guo2023TNNLS}
D.~Guo, C.~Wang, B.~Wang, and H.~Zha, ``Learning fair representations via
  distance correlation minimization,'' \emph{IEEE Transactions on Neural
  Networks and Learning Systems}, pp. 1--14, 2022.

\bibitem{Kim_2019_CVPR}
B.~Kim, H.~Kim, K.~Kim, S.~Kim, and J.~Kim, ``Learning not to learn: Training
  deep neural networks with biased data,'' in \emph{Proceedings of the IEEE/CVF
  Conference on Computer Vision and Pattern Recognition (CVPR)}, 2019, pp.
  9004--9012.

\bibitem{madras2018learning}
D.~Madras, E.~Creager, T.~Pitassi, and R.~Zemel, ``Learning adversarially fair
  and transferable representations,'' in \emph{International Conference on
  Machine Learning}.\hskip 1em plus 0.5em minus 0.4em\relax PMLR, 2018, pp.
  3384--3393.

\bibitem{Zhu_2021_ICCV}
W.~Zhu, H.~Zheng, H.~Liao, W.~Li, and J.~Luo, ``Learning bias-invariant
  representation by cross-sample mutual information minimization,'' in
  \emph{Proceedings of the IEEE/CVF International Conference on Computer Vision
  (ICCV)}, 2021, pp. 15\,002--15\,012.

\bibitem{zafar2019fairness}
E.~Creager, D.~Madras, J.-H. Jacobsen, M.~Weis, K.~Swersky, T.~Pitassi, and
  R.~Zemel, ``Flexibly fair representation learning by disentanglement,'' in
  \emph{Proceedings of the 36th International Conference on Machine Learning},
  2019, pp. 1436--1445.

\bibitem{Park2021LearningDR}
S.~Park, S.~Hwang, D.~Kim, and H.~Byun, ``Learning disentangled representation
  for fair facial attribute classification via fairness-aware information
  alignment,'' in \emph{Proceedings of the AAAI Conference on Artificial
  Intelligence}, 2021, pp. 2403--2411.

\bibitem{mroueh2021fair}
Y.~M. Ching-Yao~Chuang, ``Fair mixup: Fairness via interpolation,'' in
  \emph{International Conference on Learning Representations}, 2021.

\bibitem{du2021fairness}
M.~Du, S.~Mukherjee, G.~Wang, R.~Tang, A.~Awadallah, and X.~Hu, ``Fairness via
  representation neutralization,'' in \emph{Advances in Neural Information
  Processing Systems}, 2021.

\bibitem{Park_2022_CVPR}
S.~Park, J.~Lee, P.~Lee, S.~Hwang, D.~Kim, and H.~Byun, ``Fair contrastive
  learning for facial attribute classification,'' in \emph{Proceedings of the
  IEEE/CVF Conference on Computer Vision and Pattern Recognition (CVPR)}, 2022,
  pp. 10\,389--10\,398.

\bibitem{Wang_2022_CVPR}
Z.~Wang, X.~Dong, H.~Xue, Z.~Zhang, W.~Chiu, T.~Wei, and K.~Ren,
  ``Fairness-aware adversarial perturbation towards bias mitigation for
  deployed deep models,'' in \emph{Proceedings of the IEEE/CVF Conference on
  Computer Vision and Pattern Recognition (CVPR)}, 2022, pp. 10\,379--10\,388.

\bibitem{zhang2023fairnessaware}
F.~Zhang, K.~Kuang, L.~Chen, Y.~Liu, C.~Wu, and J.~Xiao, ``Fairness-aware
  contrastive learning with partially annotated sensitive attributes,'' in
  \emph{The Eleventh International Conference on Learning Representations},
  2023.

\bibitem{qi2022fairvfl}
T.~Qi, F.~Wu, C.~Wu, L.~Lyu, T.~Xu, H.~Liao, Z.~Yang, Y.~Huang, and X.~Xie,
  ``Fair{VFL}: A fair vertical federated learning framework with contrastive
  adversarial learning,'' in \emph{Advances in Neural Information Processing
  Systems}, 2022.

\bibitem{hwang2020UnsupervisedImagetoImageTranslation}
S.~Hwang and H.~Byun, ``\BIBforeignlanguage{en}{Unsupervised
  {{Image}}-to-{{Image Translation Via Fair Representation}} of {{Gender
  Bias}}},'' in \emph{\BIBforeignlanguage{en}{{{ICASSP}} 2020 - 2020 {{IEEE
  International Conference}} on {{Acoustics}}, {{Speech}} and {{Signal
  Processing}} ({{ICASSP}})}}.\hskip 1em plus 0.5em minus 0.4em\relax
  {Barcelona, Spain}: {IEEE}, May 2020, pp. 1953--1957.

\bibitem{joo2020GenderSlopesCounterfactual}
J.~Joo and K.~K{\"a}rkk{\"a}inen, ``Gender slopes: Counterfactual fairness for
  computer vision models by attribute manipulation,'' in \emph{Proceedings of
  the 2nd International Workshop on Fairness, Accountability, Transparency and
  Ethics in Multimedia}, 2020, pp. 1--5.

\bibitem{ramaswamy2021fair}
V.~V. Ramaswamy, S.~S. Kim, and O.~Russakovsky, ``Fair attribute classification
  through latent space de-biasing,'' in \emph{Proceedings of the IEEE/CVF
  Conference on Computer Vision and Pattern Recognition}, 2021, pp. 9301--9310.

\bibitem{roh2021fairbatch}
Y.~Roh, K.~Lee, S.~E. Whang, and C.~Suh, ``Fairbatch: Batch selection for model
  fairness,'' in \emph{International Conference on Learning Representations},
  2021.

\bibitem{Khalili2021FairSS}
M.~M. Khalili, X.~Zhang, and M.~Abroshan, ``Fair sequential selection using
  supervised learning models,'' \emph{ArXiv}, vol. abs/2110.13986, 2021.

\bibitem{zhao2020maintaining}
B.~Zhao, X.~Xiao, G.~Gan, B.~Zhang, and S.-T. Xia, ``Maintaining discrimination
  and fairness in class incremental learning,'' in \emph{Proceedings of the
  IEEE/CVF Conference on Computer Vision and Pattern Recognition}, 2020, pp.
  13\,208--13\,217.

\bibitem{gong2021mitigating}
S.~Gong, X.~Liu, and A.~K. Jain, ``Mitigating face recognition bias via group
  adaptive classifier,'' in \emph{Proceedings of the IEEE/CVF Conference on
  Computer Vision and Pattern Recognition}, 2021, pp. 3414--3424.

\bibitem{Zhang_2020}
\BIBentryALTinterwordspacing
T.~Zhang, t.~zhu, J.~Li, M.~Han, W.~Zhou, and P.~Yu, ``Fairness in
  semi-supervised learning: Unlabeled data help to reduce discrimination,''
  \emph{IEEE Transactions on Knowledge and Data Engineering}, pp. 1--1, 2020.
  [Online]. Available: \url{http://dx.doi.org/10.1109/TKDE.2020.3002567}
\BIBentrySTDinterwordspacing

\bibitem{wei2023balanced}
X.-X. Wei and H.~Huang, ``Balanced federated semisupervised learning with
  fairness-aware pseudo-labeling,'' \emph{IEEE Transactions on Neural Networks
  and Learning Systems}, 2023.

\bibitem{huanmulti24}
H.~Tian, B.~Liu, T.~Zhu, W.~Zhou, and P.~S. Yu, ``Multifair: Model fairness
  with multiple sensitive attributes,'' \emph{IEEE Transactions on Neural
  Networks and Learning Systems}, pp. 1--14, 2024.

\bibitem{Huan2024multi}
------, ``Multifair: Model fairness with multiple sensitive attributes,''
  \emph{IEEE Transactions on Neural Networks and Learning Systems}, pp. 1--14,
  2024.

\bibitem{chai2022self}
J.~Chai and X.~Wang, ``Self-supervised fair representation learning without
  demographics,'' \emph{Advances in Neural Information Processing Systems},
  vol.~35, pp. 27\,100--27\,113, 2022.

\bibitem{salemMLLeaksModelData2019}
A.~Salem, Y.~Zhang, M.~Humbert, P.~Berrang, M.~Fritz, and M.~Backes,
  ``{{ML-Leaks}}: {{Model}} and data independent membership inference attacks
  and defenses on machine learning models,'' in \emph{26th Annual Network and
  Distributed System Security Symposium, {{NDSS}} 2019}, 2019.

\bibitem{yeomPrivacyRiskMachine2018}
S.~Yeom, I.~Giacomelli, M.~Fredrikson, and S.~Jha, ``Privacy {{Risk}} in
  {{Machine Learning}}: {{Analyzing}} the {{Connection}} to {{Overfitting}},''
  in \emph{2018 {{IEEE}} 31st {{Computer Security Foundations Symposium}}
  ({{CSF}})}, 2018, pp. 268--282.

\bibitem{sablayrollesWhiteboxVsBlackbox2019}
A.~Sablayrolles, M.~Douze, C.~Schmid, Y.~Ollivier, and H.~Jegou, ``White-box vs
  {{Black-box}}: {{Bayes Optimal Strategies}} for {{Membership Inference}},''
  in \emph{International {{Conference}} on {{Machine Learning}}}, 2019, pp.
  5558--5567.

\bibitem{Liu23tdsc}
L.~Liu, Y.~Wang, G.~Liu, K.~Peng, and C.~Wang, ``Membership inference attacks
  against machine learning models via prediction sensitivity,'' \emph{IEEE
  Transactions on Dependable and Secure Computing}, vol.~20, no.~3, pp.
  2341--2347, 2023.

\bibitem{choquette-chooLabelOnlyMembershipInference2021}
C.~A. {Choquette-Choo}, F.~Tramer, N.~Carlini, and N.~Papernot, ``Label-{{Only
  Membership Inference Attacks}},'' in \emph{International {{Conference}} on
  {{Machine Learning}}}, 2021, pp. 1964--1974.

\bibitem{liMembershipLeakageLabelOnly2021}
Z.~Li and Y.~Zhang, ``Membership {{Leakage}} in {{Label-Only Exposures}},'' in
  \emph{{{ACM SIGSAC Conference}} on {{Computer}} and {{Communications
  Security}} ({{CCS}} 2021)}, 2021.

\bibitem{ye2022enhanced}
J.~Ye, A.~Maddi, S.~K. Murakonda, V.~Bindschaedler, and R.~Shokri, ``Enhanced
  membership inference attacks against machine learning models,'' in
  \emph{Proceedings of the 2022 ACM SIGSAC Conference on Computer and
  Communications Security}, 2022, pp. 3093--3106.

\bibitem{liuEncoderMIMembershipInference2021}
H.~Liu, J.~Jia, W.~Qu, and N.~Z. Gong, ``{{EncoderMI}}: {{Membership
  Inference}} against {{Pre-trained Encoders}} in {{Contrastive Learning}},''
  in \emph{Proceedings of the 2021 {{ACM SIGSAC Conference}} on {{Computer}}
  and {{Communications Security}}}, 2021, pp. 2081--2095.

\bibitem{gaoSimilarityDistributionBased2023}
J.~Gao, X.~Jiang, H.~Zhang, Y.~Yang, S.~Dou, D.~Li, D.~Miao, C.~Deng, and
  C.~Zhao, ``Similarity {{Distribution Based Membership Inference Attack}} on
  {{Person Re-identification}},'' in \emph{Proceedings of the AAAI Conference
  on Artificial Intelligence}, 2023, pp. 14\,820--14\,828.

\bibitem{yuanMembershipInferenceAttacks2022}
X.~Yuan and L.~Zhang, ``Membership {{Inference Attacks}} and {{Defenses}} in
  {{Neural Network Pruning}},'' in \emph{31st {{USENIX Security Symposium}}
  ({{USENIX Security}} 22)}, 2022, pp. 4561--4578.

\bibitem{zhang23tdsc}
G.~Zhang, B.~Liu, T.~Zhu, M.~Ding, and W.~Zhou, ``Label-only membership
  inference attacks and defenses in semantic segmentation models,'' \emph{IEEE
  Transactions on Dependable and Secure Computing}, vol.~20, no.~2, pp.
  1435--1449, 2023.

\bibitem{chenRelaxLossDefendingMembership2022}
D.~Chen, N.~Yu, and M.~Fritz, ``{{RelaxLoss}}: {{Defending Membership Inference
  Attacks}} without {{Losing Utility}},'' in \emph{International {{Conference}}
  on {{Learning Representations}}}, 2022.

\bibitem{yangPurifierDefendingData2023}
Z.~Yang, L.~Wang, D.~Yang, J.~Wan, Z.~Zhao, E.-C. Chang, F.~Zhang, and K.~Ren,
  ``Purifier: {{Defending Data Inference Attacks}} via {{Transforming
  Confidence Scores}},'' in \emph{Proceedings of the AAAI Conference on
  Artificial Intelligence}, 2023, pp. 10\,871--10\,879.

\bibitem{huang22tdsc}
H.~Huang, W.~Luo, G.~Zeng, J.~Weng, Y.~Zhang, and A.~Yang, ``Damia: Leveraging
  domain adaptation as a defense against membership inference attacks,''
  \emph{IEEE Transactions on Dependable and Secure Computing}, vol.~19, no.~5,
  pp. 3183--3199, 2022.

\bibitem{liu23tdscdefense}
Y.~Liu, H.~Li, G.~Huang, and W.~Hua, ``Opupo: Defending against membership
  inference attacks with order-preserving and utility-preserving obfuscation,''
  \emph{IEEE Transactions on Dependable and Secure Computing}, vol.~20, no.~6,
  pp. 4690--4701, 2023.

\bibitem{hu23tdscdef}
L.~Hu, J.~Li, G.~Lin, S.~Peng, Z.~Zhang, Y.~Zhang, and C.~Dong, ``Defending
  against membership inference attacks with high utility by gan,'' \emph{IEEE
  Transactions on Dependable and Secure Computing}, vol.~20, no.~3, pp.
  2144--2157, 2023.

\bibitem{heSemiLeakMembershipInference2022}
X.~He, H.~Liu, N.~Z. Gong, and Y.~Zhang, ``Semi-{{Leak}}: {{Membership
  Inference Attacks Against Semi-supervised Learning}},'' in \emph{Computer
  {{Vision}} \textendash{} {{ECCV}} 2022}, 2022, pp. 365--381.

\bibitem{liAuditingMembershipLeakages2022}
Z.~Li, Y.~Liu, X.~He, N.~Yu, M.~Backes, and Y.~Zhang, ``Auditing {{Membership
  Leakages}} of {{Multi-Exit Networks}},'' in \emph{Proceedings of the 2022
  {{ACM SIGSAC Conference}} on {{Computer}} and {{Communications Security}}},
  2022, pp. 1917--1931.

\bibitem{hu2022m}
P.~Hu, Z.~Wang, R.~Sun, H.~Wang, and M.~Xue, ``M4i: Multi-modal models
  membership inference,'' in \emph{Advances in Neural Information Processing
  Systems}, 2022, pp. 1867--1882.

\bibitem{yeom2018privacy}
S.~Yeom, I.~Giacomelli, M.~Fredrikson, and S.~Jha, ``Privacy risk in machine
  learning: Analyzing the connection to overfitting,'' in \emph{2018 IEEE 31st
  Computer Security Foundations Symposium (CSF)}.\hskip 1em plus 0.5em minus
  0.4em\relax IEEE, 2018, pp. 268--282.

\bibitem{ganju2018property}
K.~Ganju, Q.~Wang, W.~Yang, C.~A. Gunter, and N.~Borisov, ``Property inference
  attacks on fully connected neural networks using permutation invariant
  representations,'' in \emph{Proceedings of the 2018 ACM SIGSAC Conference on
  Computer and Communications Security}, 2018, pp. 619--632.

\bibitem{xu2021robust}
H.~Xu, X.~Liu, Y.~Li, A.~Jain, and J.~Tang, ``To be robust or to be fair:
  Towards fairness in adversarial training,'' in \emph{International conference
  on machine learning}.\hskip 1em plus 0.5em minus 0.4em\relax PMLR, 2021, pp.
  11\,492--11\,501.

\bibitem{zhang2023revisiting}
T.~Zhang, T.~Zhu, J.~Li, W.~Zhou, and S.~Y. Philip, ``Revisiting model fairness
  via adversarial examples,'' \emph{Knowledge-Based Systems}, p. 110777, 2023.

\bibitem{Chang2020OnAB}
H.~Chang, T.~D. Nguyen, S.~K. Murakonda, E.~Kazemi, and R.~Shokri, ``On
  adversarial bias and the robustness of fair machine learning,'' \emph{ArXiv},
  vol. abs/2006.08669, 2020.

\bibitem{ijcai2023p59}
H.~Zeng, Z.~Yue, L.~Shang, Y.~Zhang, and D.~Wang, ``On adversarial robustness
  of demographic fairness in face attribute recognition,'' in \emph{Proceedings
  of the Thirty-Second International Joint Conference on Artificial
  Intelligence, {IJCAI-23}}, 2023, pp. 527--535.

\bibitem{mehrabi2021exacerbating}
N.~Mehrabi, M.~Naveed, F.~Morstatter, and A.~Galstyan, ``Exacerbating
  algorithmic bias through fairness attacks,'' in \emph{Proceedings of the AAAI
  Conference on Artificial Intelligence}, vol.~35, no.~10, 2021, pp.
  8930--8938.

\bibitem{aalmoes2022leveraging}
J.~Aalmoes, V.~Duddu, and A.~Boutet, ``On the alignment of group fairness with
  attribute privacy,'' 2024.

\bibitem{balunovic2022fair}
\BIBentryALTinterwordspacing
M.~Balunovic, A.~Ruoss, and M.~Vechev, ``Fair normalizing flows,'' in
  \emph{International Conference on Learning Representations}, 2022. [Online].
  Available: \url{https://openreview.net/forum?id=BrFIKuxrZE}
\BIBentrySTDinterwordspacing

\bibitem{zhao-etal-2017-men}
J.~Zhao, T.~Wang, M.~Yatskar, V.~Ordonez, and K.-W. Chang, ``Men also like
  shopping: Reducing gender bias amplification using corpus-level
  constraints,'' in \emph{Proceedings of the 2017 Conference on Empirical
  Methods in Natural Language Processing}, 2017, pp. 2979--2989.

\bibitem{Hardt:2016wv}
M.~Hardt, E.~Price, and N.~Srebro, ``{Equality of opportunity in supervised
  learning},'' in \emph{Advances in neural information processing systems},
  2016, pp. 3315--3323.

\bibitem{CelebAMask-HQ}
C.-H. Lee, Z.~Liu, L.~Wu, and P.~Luo, ``Maskgan: Towards diverse and
  interactive facial image manipulation,'' in \emph{Proceedings of the IEEE/CVF
  Conference on Computer Vision and Pattern Recognition (CVPR)}, 2020.

\bibitem{geraldsutkface}
J.~Geralds, ``Utkface large scale face dataset,'' \emph{github. com}, 2017.

\bibitem{karkkainen2021fairface}
K.~Karkkainen and J.~Joo, ``Fairface: Face attribute dataset for balanced race,
  gender, and age for bias measurement and mitigation,'' in \emph{Proceedings
  of the IEEE/CVF Winter Conference on Applications of Computer Vision}, 2021,
  pp. 1548--1558.

\bibitem{zhang2023fairness}
F.~Zhang, K.~Kuang, L.~Chen, Y.~Liu, C.~Wu, and J.~Xiao, ``Fairness-aware
  contrastive learning with partially annotated sensitive attributes,'' in
  \emph{The Eleventh International Conference on Learning Representations},
  2023.

\bibitem{PinzonPPV22}
C.~Pinz{\'{o}}n, C.~Palamidessi, P.~Piantanida, and F.~Valencia, ``On the
  impossibility of non-trivial accuracy in presence of fairness constraints,''
  in \emph{Thirty-Sixth {AAAI} Conference on Artificial Intelligence}, 2022,
  pp. 7993--8000.

\bibitem{Zietlow_2022_CVPR}
D.~Zietlow, M.~Lohaus, G.~Balakrishnan, M.~Kleindessner, F.~Locatello,
  B.~Sch\"olkopf, and C.~Russell, ``Leveling down in computer vision: Pareto
  inefficiencies in fair deep classifiers,'' in \emph{Proceedings of the
  IEEE/CVF Conference on Computer Vision and Pattern Recognition (CVPR)}, 2022,
  pp. 10\,410--10\,421.

\bibitem{wangMitigatingBiasFace}
M.~Wang and W.~Deng, ``Mitigating {{Bias}} in {{Face Recognition Using
  Skewness}}-{{Aware Reinforcement Learning}},'' in \emph{Proceedings of the
  IEEE/CVF Conference on Computer Vision and Pattern Recognition (CVPR)}, 2020,
  p.~10.

\bibitem{han2023ffb}
X.~Han, J.~Chi, Y.~Chen, Q.~Wang, H.~Zhao, N.~Zou, and X.~Hu, ``Ffb: A fair
  fairness benchmark for in-processing group fairness methods,'' 2023.

\bibitem{He2016DeepRL}
K.~He, X.~Zhang, S.~Ren, and J.~Sun, ``Deep residual learning for image
  recognition,'' \emph{Proceedings of the IEEE/CVF Conference on Computer
  Vision and Pattern Recognition (CVPR)}, pp. 770--778, 2016.

\bibitem{Simonyan2015VeryDC}
K.~Simonyan and A.~Zisserman, ``Very deep convolutional networks for
  large-scale image recognition,'' \emph{CoRR}, vol. abs/1409.1556, 2015.

\bibitem{dwork2006calibrating}
C.~Dwork, F.~McSherry, K.~Nissim, and A.~Smith, ``Calibrating noise to
  sensitivity in private data analysis,'' in \emph{Theory of Cryptography:
  Third Theory of Cryptography Conference}, 2006, pp. 265--284.

\bibitem{abadi2016deep}
M.~Abadi, A.~Chu, I.~Goodfellow, H.~B. McMahan, I.~Mironov, K.~Talwar, and
  L.~Zhang, ``Deep learning with differential privacy,'' in \emph{Proceedings
  of the 2016 ACM SIGSAC conference on computer and communications security},
  2016, pp. 308--318.

\bibitem{zhang2024unraveling}
H.~Zhang, X.~Yuan, and S.~Pan, ``Unraveling privacy risks of individual
  fairness in graph neural networks,'' in \emph{2024 IEEE 40th International
  Conference on Data Engineering (ICDE)}.\hskip 1em plus 0.5em minus
  0.4em\relax IEEE, 2024, pp. 1712--1725.

\end{thebibliography}
}

\end{document}